\documentclass[]{fairmeta}

\usepackage[utf8]{inputenc}

\usepackage{amsmath}
\usepackage{amssymb}

\usepackage{tabularx}
\usepackage{colortbl}
\usepackage{wrapfig}

\usepackage{adjustbox}
\usepackage{placeins}

\definecolor{colorNoAfd}{HTML}{008080}
\definecolor{colorPi0}{HTML}{A2CD5A}
\definecolor{colorOursScratch}{HTML}{6495ED}
\definecolor{colorOursPretrained}{HTML}{FF6347}
\definecolor{pale_pink}{rgb}{0.98, 0.9, 0.95}

\newtcolorbox{findingbox}{%
  colback=metabg,
  colframe=metablue,
  arc=2mm,
  boxrule=0.8pt,
  left=6pt,right=6pt,top=4pt,bottom=4pt,
  enhanced,
  breakable
}

\newtcolorbox{promptbox}[1]{%
  colback=metabg,
  colframe=metablue,
  arc=1mm,
  boxrule=0.6pt,
  left=5pt,right=5pt,top=3pt,bottom=3pt,
  enhanced,
  breakable,
  fonttitle=\bfseries\small,
  coltitle=white,
  fontupper=\small,
  title={#1}
}

\usepackage{xspace}
\providecommand{\eg}{\emph{e.g.}\xspace}
\providecommand{\ie}{\emph{i.e.}\xspace}
\providecommand{\cf}{\emph{cf.}\xspace}

\providecommand{\keywords}[1]{}

\title{AffordanceVLA: A Vision-Language-Action Model Empowering Action Generation through Affordance-Aware Understanding}

\author[1,\dagger]{Qize Yu}
\author[2,\dagger]{Jiadi You}
\author[1]{Yuran Wang}
\author[1]{Jiaqi Liang}
\author[1]{Bowen Ping}
\author[1]{Yang Tian}
\author[1]{Yue Chen}
\author[1,3]{Bohan Zhou}
\author[2]{Wenxuan Song}
\author[3]{Minghong Cai}
\author[2]{Zeying Gong}
\author[1]{Ruihai Wu}
\author[4]{Yinchuan Li}
\author[2,*]{Junwei Liang}
\author[2,*]{Yingcong Chen}

\affiliation[1]{Peking University}
\affiliation[2]{HKUST (GZ)}
\affiliation[3]{The Chinese University of Hong Kong}
\affiliation[4]{Knowin AI}

\contribution[\dagger]{Equal contribution.\quad${}^{*}$Corresponding authors.}

\abstract{
Vision-Language-Action (VLA) models leverage the rich world knowledge of pretrained vision-language models (VLMs) to enable instruction-following robotic manipulation.
However, the structural mismatch between VLM semantic spaces and embodied control policies often hinders the learning of precise perception--action mappings.
To address this challenge, we propose AffordanceVLA, a scalable, unified framework that introduces structured affordance learning as a task-oriented and object-centric intermediate representation to establish a more precise and robust perception--action mapping.
Specifically, we progressively model manipulation priors through three complementary components—Which2Act, Where2Act, and How2Act—which jointly produce spatially grounded, semantically conditioned, and action-coupled intermediate representations, thereby naturally bridging vision, language, and action.
We integrate these modules into a Mixture-of-Transformer (MoT) architecture and adopt a three-stage progressive training curriculum that leverages 8.8M affordance-centric grounding and pointing samples, translating scalable vision-language supervision into improved robotic manipulation.
To overcome the scarcity of dense affordance labels in robotic datasets, we also develop a robust automated data augmentation pipeline.
Extensive experiments on ID \& OOD simulation and real-world demonstrate that AffordanceVLA outperforms strong VLA baselines across diverse manipulation scenarios, validating structured affordance learning as a scalable and robust bridge that integrates multi-source vision-language supervision while connecting pretrained VLM semantics to embodied robot control.
}

\metadata[Project Page]{\url{https://skywalker-yqz.github.io/AffordanceVLA/}}
\metadata[Code]{\url{https://github.com/Skywalker-yqz/AffordanceVLA}}

\begin{document}

\maketitle

\section{Introduction}
\label{sec:intro}

\begin{figure}[!t]
    \centering
    \includegraphics[width=\linewidth]{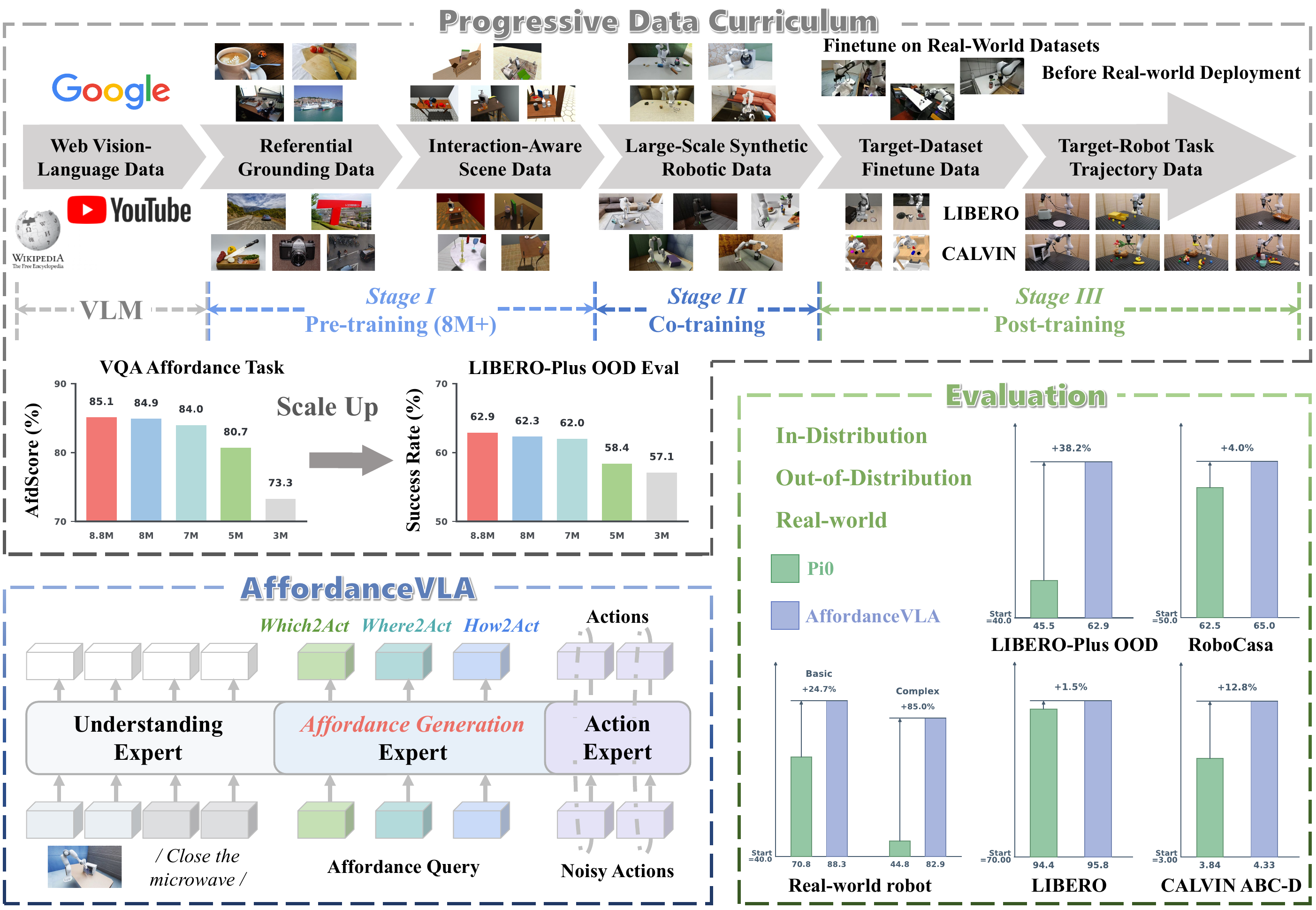}
    \caption{\textbf{AffordanceVLA Overview.} 
    \textbf{(Bottom-left)} AffordanceVLA employs three specialized experts (Understanding, Affordance Generation, and Action), leveraging structured affordance forecasting (Which2Act, Where2Act, and How2Act) as intermediate representations to bridge perception and action. \textbf{(Top)} Enabled by a three-stage training strategy with progressive data curriculum, \textbf{(Bottom-right)} AffordanceVLA achieves strong performance.
    }
    \label{fig:overview}
\end{figure}

Vision-Language-Action (VLA) models~\cite{openvla, octo, pi0, diffusionvla, rdt, rmbench} have shown great promise in instruction-following manipulation, largely driven by the rich world knowledge embedded in pretrained Vision-Language Models (VLMs). Many approaches~\cite{cogact, tinyvla, spatialvla, gr00tn1, smolvla, geminirobotics, instructvla} attempt a direct end-to-end mapping from natural language instructions and visual observations to robot actions. However, while the core of VLM pre-training lies in aligning vision and language within a semantic space, robotic actions are inherently representations in the 3D physical space, creating a significant gap that makes learning a direct mapping challenging. Therefore, finding an appropriate intermediate representation to bridge perception and action is crucial.

Previous works have made substantial efforts to bridge perception and action. A conventional paradigm employs video prediction or visual foresight~\cite{seer, upvla, cotvla, unifiedworldmodels, reinbot, f1vla, wang2026openwamopenmodularexploration, song2026uniwamunifiedworldactionmodel} to guide action generation. However, such dense visual signals often contain redundant information, and the inference process of video prediction models is typically time-consuming. 
In addition to video prediction, many works incorporate visual representations such as depth, visual grounding~\cite{yu2026groundanythingreconcilingparalleldecoding}, and optical flow~\cite{bridgevla, 3dvla, worldvla, motus} to enhance visual localization.
Overall, the exploration and utilization of intermediate representations remains an open question.

We argue that a robust VLA requires learning a richer, task-oriented representation to enhance generalization across complex manipulation scenarios. Without innovating the model paradigm, blindly scaling up data fails to maximize the intrinsic power within the datasets, and relying solely on scaling is insufficient to resolve the fundamental spatial gap. Indeed, recent strong VLAs increasingly introduce structured, train-only intermediate supervision to better exploit their data and to keep the backbone's vision--language ability from being eroded by low-level action learning~\cite{ye2026augustusautonomousdrivingvla}---a direction we pursue through affordance.

\begin{findingbox}
Without innovating the model paradigm, blindly scaling up data fails to maximize the intrinsic power within the datasets, and relying solely on scaling is insufficient to resolve the fundamental spatial gap.
\end{findingbox}

Intuitively, much like how humans naturally perceive a mug's handle as an invitation to grasp, an embodied agent should understand its environment. \textbf{Affordances}~\cite{gibson1977theory}—manipulation priors that explicitly indicate \textit{which} object to manipulate, as well as \textit{where} and \textit{how} to interact—naturally fulfill this role as ideal intermediate prediction targets~\cite{afforddp, coavla, pa3ff, eqvafford, you2026affordancewamaffordanceawarejointworldaction}. They serve as a perfect bridge by seamlessly coupling spatial grounding in vision, semantic conditioning in language, and execution guidance in action. Furthermore, extensive prior work~\cite{vatmart, dexgarmentlab, A3D, garmentpile, where2explore, adamanip, ram, dualafford} has shown that affordance representations are readily learnable and highly adaptable.

\begin{findingbox}
Affordances serve as a perfect bridge, seamlessly coupling spatial grounding in vision, semantic conditioning in language, and execution guidance in action.
\end{findingbox}

Building upon these insights, we propose \textbf{AffordanceVLA} (as illustrated in \Cref{fig:overview}), a novel framework that leverages affordance forecasting to establish a precise and robust perception-action mapping. To comprehensively integrate this task-relevant information, our framework incorporates three streamlined designs: \textbf{Which2Act} achieves object-centric grounding by predicting the visual latent of the target to suppress irrelevant distractions; \textbf{Where2Act} predicts a 2D affordance map to pinpoint precise interaction areas; and \textbf{How2Act} extracts a 3D geometric representation to guide the manipulation policy. Together, these modules progressively weave affordance priors—from coarse to fine, and from 2D to 3D—into the VLA's perception-action loop, equipping the model with a structured and deeply grounded understanding of the physical world.

To effectively instantiate these structured affordance designs and bridge the modality gap, we build AffordanceVLA upon a Mixture-of-Transformer (MoT)~\cite{mot} architecture comprising three dedicated experts: an \textit{Understanding Expert}, an \textit{Affordance Generation Expert}, and an \textit{Action Expert}. This decoupled multi-expert design facilitates progressive information fusion and representation propagation from broad perception to closed-loop control. Moreover, the architecture naturally absorbs diverse multimodal knowledge from varied sources, such as VQA and robotic trajectories. Crucially, since structured affordance is expressed in a VQA-compatible form, it bridges not only vision--language perception and action, but also the two data ecosystems behind them: web-scale multimodal corpora and scarce robot trajectories. We therefore curate \textbf{8.8M} affordance-centric VQA samples---collected from real annotations rather than synthesized---and our scaling study shows that manipulation performance improves consistently as this supervision grows, supporting our central claim:

\begin{findingbox}
Scalable VQA affordance supervision $\rightarrow$ affordance–action co-training $\rightarrow$ Scalable robot control.
\end{findingbox}

Recognizing the inherent lack of dense affordance annotations in large-scale robotic datasets, we further develop a robust data augmentation pipeline to synthesize these critical supervisory signals for co-training. Supported by this tool, we introduce a three-stage training strategy with a progressive data curriculum: Stage I Pre-training on referential grounding and interaction-aware scene data; Stage II Co-training with large-scale synthetic robotic data; and Stage III Post-training via target-dataset fine-tuning.
AffordanceVLA strongly improves its Pi0 backbone across ID \& OOD simulation benchmarks---LIBERO~\cite{libero}(+1.5\%), CALVIN~\cite{calvin}(+12.8\%), RoboCasa~\cite{nasiriany2024robocasalargescalesimulationeveryday}(+4.0\%), LIBERO-Plus~\cite{fei2025liberoplusindepthrobustnessanalysis}(+38.2\%) and real-world experiments(+24.7\%, +85.0\%), achieving competitive performance and generalization with recent strong VLAs.

In summary, our key contributions are summarized as follows:

\begin{itemize}
    \item We propose AffordanceVLA, a perception–prediction–action VLA framework that leverages structured affordance forecasting as intermediate supervision. By modeling Which2Act, Where2Act, and How2Act, our method integrates object grounding, 2D interaction localization, and 3D geometric reasoning to establish a more precise perception–action mapping.

    \item We show that such a representation is intrinsically scalable: as affordance forecasting reduces to a VQA-style prediction task, it inherits the scalability of multimodal data. Trained on 8.8M affordance-centric VQA samples, our scaling experiments reveal a consistent transfer from scalable VQA supervision to scalable robot control.

    \item We design a MoT architecture with specialized experts and a three-stage training strategy with progressive data curriculum. Affordance here acts as a bridge not only between vision--language and action, but also between massive VQA corpora and limited robotic data, enabling a transition from VQA task to embodied manipulation. Furthermore, we develop a novel robotic data augmentation pipeline to overcome the lack of affordance annotations for co-training robot data.

    \item AffordanceVLA strongly improves its Pi0 backbone across ID \& OOD simulation benchmarks and real-world experiments, achieving competitive performance and generalization with recent strong VLAs and its OOD language-conditioned performance further validate our semantic-anchor hypothesis.
\end{itemize}

\section{Related Work}
\label{sec:related_work}

\subsection{Vision-Language-Action Models}


The rapid evolution of Multimodal Large Language Models (MLLMs)~\cite{visualinstructiontuning, gpt4, qwen3} has catalyzed the development of Vision--Language--Action (VLA) models, which couple a vision--language backbone with an action module to inherit web-scale visual--linguistic priors for flexible instruction following~\cite{octo, openvla, openvla_oft, pi0, rdt, diffusionvla, eo1}. 
These models differ chiefly in how actions are represented and decoded: some autoregressively emit discretized action tokens~\cite{openvla, gr3}, whereas others attach a continuous diffusion or flow-matching action expert for smoother, high-frequency control~\cite{pi0, pi05, rdt, cogact, smolvla}. 
To improve transfer and scalability, recent efforts pursue cross-embodiment training with latent action spaces~\cite{xvla, univla}, spatially enhanced backbones~\cite{spatialvla, yu2026groundingpigroundingfoundationmodel}, and efficient lightweight variants~\cite{tinyvla, smolvla}. 
Beyond monolithic policies, dual-system and expert-decoupled designs further disentangle high-level understanding from reactive control~\cite{openhelix, gr00tn1, geminirobotics, zhou2026roboawarelearningcoordinateembodied}. 

Despite their strong perceptual grounding, directly regressing actions from raw observations leaves VLAs without explicit reasoning about task-relevant scene structure, which often undermines robustness. 
A large body of work therefore augments VLAs with auxiliary intermediate representations, broadly falling into two families. 
The first treats future prediction as a \emph{world model}, using generative visual foresight---video or image prediction~\cite{LUP, gr-1, susie, vpp, clover, cotvla, super-mimic}, predictive inverse dynamics~\cite{seer}, and unified video--action or latent-action world models~\cite{unifiedworldmodels, worldvla, motus, 3dvla, dreamvla, f1vla, upvla, reinbot, flowvla, shen2026ld4wamlearninglatentdynamics}---to guide action generation. 
While visually intuitive, such dense pixel-level targets are highly redundant and their rollout is typically computationally heavy. 
The second family instead predicts \emph{more compact structured cues}, such as textual rationales or chains of thought~\cite{instructvla, hirobot, thinkact}, keyposes and object pointflow~\cite{gripperkeypose}, reconstructive or gaze-region targets~\cite{reconvla}, and implicit 3D/spatial representation alignment~\cite{spatialforcing}. 
Notably, the latest generalist policies (\eg $\pi_{0.5}$~\cite{pi05} and $\pi_{0.7}$~\cite{pi07}) reveal a complementary trend: introducing train-only structured intermediate supervision (\eg discrete action tokens or bounding boxes that are \emph{not} deployed at inference), so that the low-level control objective does not erode the backbone's vision--language and instruction-following ability. 

These intermediates, however, are frequently either overly redundant (dense video) or too coarse (global rationales) to capture the task-critical \textit{what}, \textit{where}, and \textit{how} of manipulation. 
In contrast, we adopt \textbf{affordance} as a structured, task-focused intermediate representation that is simultaneously spatially grounded, semantically conditioned, and action-coupled---distinct from both video-prediction-as-bridge and latent-action world models. 
By employing a multi-stage training strategy, our approach integrates deep semantic understanding with world knowledge, leading to a more reliable perception--action mapping and enhanced robustness.

\subsection{Affordance for Robotic Manipulation}

Affordance~\cite{gibson1977theory} characterizes the actionable possibilities that objects offer, providing a compact prior for \textit{what} can be manipulated and \textit{how} to interact with it. 
It has been extensively studied across robotic grasping~\cite{ganhand, learninggrasping, roboticpnp, anygrasp}, articulated-object manipulation~\cite{vatmart, where2act, where2explore, adamanip, generalflow, actthepart, pa3ff, eqvafford, A3D}, deformable-object manipulation~\cite{dexgarmentlab, garmentpile, garmentpile2, etseed}, and broader scene interaction~\cite{interaction-exploration, ego-topo, hassanin2021visual}, consistently demonstrating strong cross-task generalization. 
Owing to its expressive power, affordance has been incorporated into diverse learning paradigms: some methods learn it from large-scale human-video datasets or reinforcement learning~\cite{afhv, endtoendaffordance, where2act, dualafford}, while others, such as Robo-ABC~\cite{roboabc} and RAM~\cite{ram}, enable training-free affordance transfer across novel objects and tasks via semantic correspondence and retrieval. 

Nevertheless, most prior work represents affordances as sparse 2D/3D contact points with directions and couples them with external grasp generators~\cite{anygrasp} and motion planners~\cite{moveit, curobo}, forming open-loop pipelines that are brittle on long-horizon or contact-rich tasks. 
While AffordDP~\cite{afforddp} and CoA-VLA~\cite{coavla} move toward closed-loop control, affordance is still largely consumed as an external cue and does not fully exploit the rich world knowledge embedded in large-scale pre-trained VLMs. 
In contrast, we internalize structured affordance forecasting \textit{inside} a web-scale pre-trained VLA, jointly optimizing it with the VLM backbone and the action expert across semantic, operational, and spatial--geometric dimensions. 
This embedding turns affordance from an externally consumed prior into an action-coupled intermediate representation, improving reasoning, generalization, and performance across diverse manipulation scenarios.

\begin{figure*}[htb]
    \centering
    \includegraphics[width=1.0\textwidth]{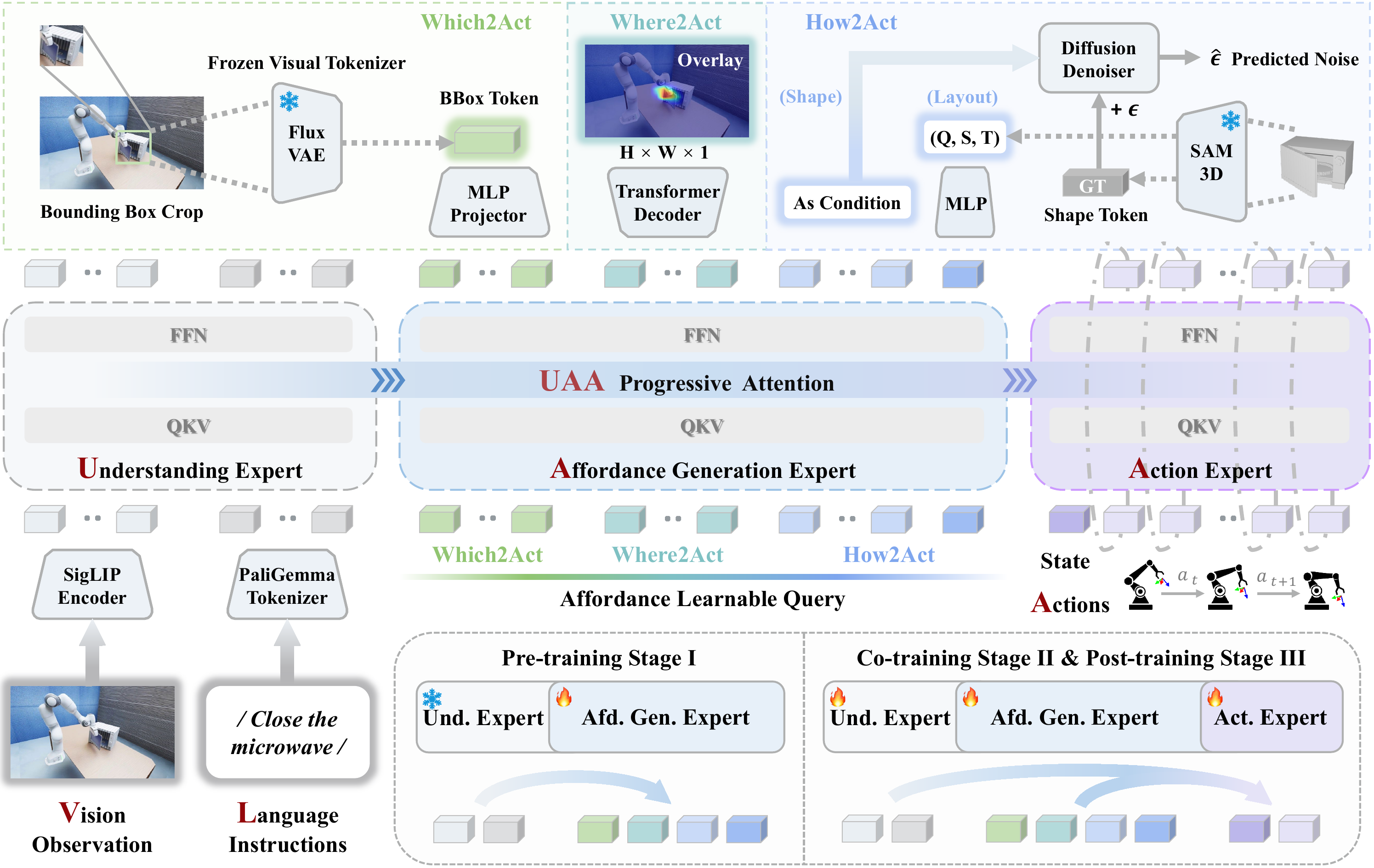}
    \caption{\textbf{Pipeline.} The framework employs a MoT architecture comprising three specialized experts of--Understanding ($\mathcal{M}_{und}$), Affordance Generation ($\mathcal{M}_{gen}$), and Action ($\mathcal{M}_{act}$), which are coordinated via a unidirectional Understanding-Affordance-Action (UAA) progressive attention mechanism. Given an RGB observation $O_t$ and instruction $l$, $\mathcal{M}_{und}$ extracts fused semantics $h_t^{und}$. $\mathcal{M}_{gen}$ then decodes $h_t^{und}$ into structured affordance tokens $\hat{A}_{t}$ (\textit{Which2Act}, \textit{Where2Act}, \textit{How2Act}) as intermediate priors. Finally, $\mathcal{M}_{act}$ synthesizes control actions $\hat{a}_{t:t+k}$ conditioned on both $h_t^{und}$ and $\hat{A}_{t}$.}
    \label{fig:pipeline}
\end{figure*}

\section{Method}
\label{sec:method}

AffordanceVLA unifies perception, prediction, and action within a Vision-Language-Action framework by leveraging structured affordance forecasting as intermediate supervision in a Mixture-of-Transformer (MoT) architecture to learn a precise, robust, and generalizable perception-action mapping for embodied manipulation. The overall pipeline is shown in \Cref{fig:pipeline}. 
In this section, we first present the architecture of AffordanceVLA together with its overall operational pipeline(\Cref{method:architecture}). We then provide a detailed description of the three affordance knowledge prediction modules(\Cref{method:affordance_prediction}). Finally, we elaborate on the training procedure of the entire framework(\Cref{method:training}).

\subsection{AffordanceVLA Architecture}
\label{method:architecture}

As illustrated in \Cref{fig:pipeline}, the core of AffordanceVLA comprises three specialized experts: Understanding Expert, Affordance Generation Expert, and Action Expert. This Mixture-of-Transformer (MoT) architecture seamlessly bridges semantic alignment, affordance-oriented state prediction, and action execution.

\subsubsection{Understanding Expert.}

The Understanding Expert ($\mathcal{M}_{und}$) establishes a fine-grained alignment between visual perception and linguistic intent by leveraging pre-trained VLM priors. 
Given the visual observation $O_t \in \mathbb{R}^{H \times W \times 3}$ and language instruction $l$ at time step $t$, $\mathcal{M}_{und}$ dynamically fuses these multimodal inputs into an instruction-aware representation: $h_t^{und} = \mathcal{M}_{und}(O_t, l)$. The proprioceptive state $s_t$ bypasses this expert and is fed directly to $\mathcal{M}_{act}$, following the standard $\pi_0$-style decoupling. 
Moving beyond global feature extraction, this module meticulously grounds linguistic concepts into spatial visual patches, providing a robust semantic foundation for subsequent physically grounded planning.

\subsubsection{Affordance Generation Expert.} 
Acting as a specialized visual planner, this module generates structured affordance priors that bridge the gap between vision, language, and action. 
Conditioned on the instruction-aware semantics $h_t^{und}$, it predicts a structured representation $\hat{A}_{t} = \mathcal{M}_{gen}(h_t^{und})$. 
Rather than a generic feature, this representation anchors high-level semantic alignment into actionable geometric cues. 
By encapsulating spatial localization (visual latent of the target), interaction hotspots (2D affordance maps), and 3D geometric structures (shape latents), the module transforms abstract reasoning into grounded physical states. 
Consequently, it filters out task-irrelevant noise and provides decoupled, highly informative planning targets to assist downstream action generation.

\subsubsection{Action Expert.}
Conditioned on the high-level semantic context $h_t^{und}$ and physically grounded affordance $\hat{A}_{t}$, the Action Expert ($\mathcal{M}_{act}$) employs a robust generative process to decode these unified representations into low-level action. 
It synthesizes a sequence of smooth, temporally coherent, and physically plausible action chunks: $\hat{a}_{t:t+k} = \mathcal{M}_{act}(h_t^{und}, \hat{A}_{t}, s_t)$. 
By conditioning on the intermediate affordance bridge, $\mathcal{M}_{act}$ is relieved from the burden of complex visual reasoning, allowing it to focus entirely on precise and robust physical execution.

\subsubsection{Attention Mechanism.}
\label{attention_mechanism}
To coordinate the heterogeneous experts efficiently, AffordanceVLA employs a structured \textit{Understanding–Affordance–Action (UAA) progressive attention} mechanism. Within each expert, bidirectional intra-expert attention is applied to ensure thorough contextual fusion. Across modules, a strict causal inter-expert attention is enforced. Specifically, the Affordance Generation Expert queries features exclusively from the Understanding Expert (i.e., $\text{Attention}(Q_{gen}, K_{und}, V_{und})$), while the Action Expert attends to the outputs of both preceding experts. This unidirectional information flow prevents action information leakage into the prediction stage, thereby maintaining the purity of affordance features and enhancing generalization in complex environments.

\subsection{Affordance Knowledge Prediction}
\label{method:affordance_prediction}
To bridge high-level semantics and low-level control, the Affordance Generation Expert predicts structured affordance knowledge rather than monolithic global features. We disentangle learnable affordance queries into three parallel sub-modules to concurrently decode manipulation priors: \textit{object-centric visual grounding (Which2Act)}, \textit{fine-grained interaction localization (Where2Act)}, and \textit{3D spatial-geometric reasoning (How2Act)}. By leveraging bidirectional attention to jointly refine their representations, this design yields task-relevant priors that naturally unify vision, language, and action.

\begin{findingbox}
Affordance is a natural vision--language--action bridge---spatially grounded, semantically conditioned, and action-coupled---that anchors the VLM's semantics while directly serving action generation.
\end{findingbox}

\subsubsection{Which2Act Forecasting.}

To align semantic intents with specific visual entities, the \textbf{Which2Act} module performs object-centric grounding by localizing latent target bounding boxes and extracting their visual representations. 
Specifically, we crop the observation based on the target bounding box and extract a continuous visual latent $z_q \in \mathbb{R}^{C \times H \times W}$ using a frozen pre-trained encoder (e.g., Flux VAE~\cite{flux}). 
The module utilizes Which2Act queries to reconstruct the predicted latent $\hat{z}$, optimizing alignment via a Mean Squared Error (MSE) loss:
\begin{equation}
    \mathcal{L}_{\text{which}} = \frac{1}{C \cdot H \cdot W} \sum_{c, h, w} \left\| \hat{z}_{c,h,w} - z_{q,c,h,w} \right\|^2
    \label{eq:which2act}
\end{equation}
This auto-encoding formulation translates spatial localization into latent reconstruction, compelling the model to isolate the interacting entity while filtering out background distractions to provide a pure visual anchor.

\subsubsection{Where2Act Forecasting.}

The \textbf{Where2Act} module achieves fine-grained interaction localization by predicting a 2D affordance map to pinpoint interactive regions. 
To transform 1D query tokens into a 2D spatial distribution, we employ a lightweight Transformer decoder that treats spatial position embeddings as queries to extract interaction cues via cross-attention. 
The resulting features are mapped into spatial logits $\hat{y} \in \mathbb{R}^{H_t \times W_t}$ and aligned with the ground-truth heatmap $M \in [0, 1]^{H_t \times W_t}$ using a pixel-wise Binary Cross-Entropy (BCE) loss:
\begin{equation}
    \mathcal{L}_{\text{where}} = -\frac{1}{H_t W_t} \sum_{i=1}^{H_t W_t} \Big[ M_i \log \sigma(\hat{y}_i) + (1 - M_i) \log \big(1 - \sigma(\hat{y}_i)\big) \Big]
    \label{eq:where2act}
\end{equation}
where $\sigma(\cdot)$ is the sigmoid function. By ``unfolding'' compressed 1D queries into intuitive 2D hotspots, this module translates semantic intent into actionable visual affordance, providing explicit contact point guidance for low-level planning.

\subsubsection{How2Act Forecasting.}

The \textbf{How2Act} module performs 3D spatial--geometric reasoning by bifurcating its tokens into two synergistic branches: 3D shape generation and spatial layout regression. 
The shape generation branch formulates 3D voxel prediction as a conditional diffusion process, where an iterative Transformer denoiser $\hat{\epsilon}_\theta$ generates the target's 3D shape latent, optimized via a standard noise-prediction objective:
\begin{equation}
    \mathcal{L}_{\text{shape}} = \mathbb{E}_{t \sim \mathcal{U}(0, T), \epsilon \sim \mathcal{N}(0, \mathbf{I})} \left[ \left\| \epsilon - \hat{\epsilon}_\theta(x_t, t, \bar{h}_{shape}) \right\|^2 \right]
    \label{eq:shape}
\end{equation}
Concurrently, the layout regression branch employs an MLP to regress a 10-DoF spatial layout vector $\hat{y}_{layout}$ (rotation, scale, and translation), optimized via a component-wise Smooth-L1 loss:
\begin{equation}
    \mathcal{L}_{\text{layout}} = \frac{1}{10} \sum_{j=1}^{10} \text{SmoothL}_{1}(\hat{y}_{layout}^{(j)}, y_{layout}^{(j)})
    \label{eq:layout}
\end{equation}
Together, these branches thoroughly characterize the target's 3D geometry and spatial posture, equipping the Action Expert with comprehensive spatial priors and kinematic constraints.




\subsection{Training Strategy}
\label{method:training}

To fully unlock the potential of the MoT architecture and effectively bridge semantic reasoning with physical execution, we employ a three-stage training strategy with a progressive data curriculum. 

\subsubsection{Stage I: General Affordance Grounding Pre-training.}


Stage I equips the Affordance Generation Expert with broad spatial grounding and interaction-localization priors using the curated \textbf{Grounding} and \textbf{Pointing} pools in \Cref{tab:stage1_data_pool}. The Grounding pool trains object-centric localization from box, region, and referring-expression supervision; the Pointing pool trains interaction-aware localization from point, heatmap, mask, grasp, and embodied supervision. We normalize native targets to shared output formats, then filter, deduplicate, and task-balance them into 8.8M training instances.

Given scarce native 3D annotations, the localized bounding boxes, masks, and interaction cues also seed our SAM-3D~\cite{sam3d} augmentation to construct aligned Shape and Layout targets. Thus, scalable 2D data provide the entry point for scaling paired 3D affordance supervision, rather than being expected to recover 3D geometry without a 3D training target.

\begin{table}[tb]
  \centering
  \scriptsize
  \renewcommand{\arraystretch}{1.06}
  \setlength{\tabcolsep}{4pt}
  \caption{\textbf{Stage-I Grounding and Pointing data pool.}
  Parenthetical values denote millions of native spatial-supervision records (M).}
  \label{tab:stage1_data_pool}

  \begin{tabularx}{\linewidth}{@{}>{\raggedright\arraybackslash}X >{\raggedright\arraybackslash}X@{}}
    \toprule
    \textbf{Grounding Pool} & \textbf{Pointing Pool} \\
    \midrule
    Objects365~\cite{Shao_2019_ICCV} (30.000M)
      & SA-1B~\cite{sam} (1,100.000M) \\
    Open Images V7~\cite{Kuznetsova_2020} (15.852M)
      & RefSpatial~\cite{RefSpatial} (2.500M) \\
    Visual Genome~\cite{krishna2016visualgenomeconnectinglanguage} (5.400M)
      & PixMo-Points~\cite{deitke2024molmopixmoopenweights} (2.380M) \\
    Flickr30K Entities~\cite{plummer2016flickr30kentitiescollectingregiontophrase} (0.276M)
      & PRISM~\cite{graspmolmo} (0.412M) \\
    RefCOCO / RefCOCO+ / RefCOCOg~\cite{nagaraja2016modelingcontextobjectsreferring} (0.388M)
      & AGD20K~\cite{AGD20K} (0.020M) \\
    LVIS v1~\cite{8954457} (2.200M)
      & ACRONYM~\cite{eppner2020acronymlargescalegraspdataset} (17.700M) \\
    MS COCO~\cite{lin2015microsoftcococommonobjects} (2.500M)
      & RoboAfford~\cite{hao2025roboaffordgenerativeaienhanceddataset} (1.900M) \\
    LLaVA-OneVision-2-Spatial~\cite{an2026llavaonevision2nextgenerationperceptualintelligence} (4.000M)
      & RoboInter-VQA / Data~\cite{li2026robointerholisticintermediaterepresentation} (2.300M) \\
    Ego4D-FHO~\cite{grauman2022ego4dworld3000hours} (0.091M)
      & RoboSpatial~\cite{song2026robospatialteachingspatialunderstanding} (3.000M) \\
    VISOR~\cite{darkhalil2022epickitchensvisorbenchmarkvideo} (9.900M)
      & HOI4D~\cite{liu2024hoi4d4degocentricdataset} (2.400M) \\
    MECCANO~\cite{ragusa2023meccano} (0.202M)
      & H2O~\cite{Kwon_2021_ICCV} (0.572M) \\
    GEM-250K(4M)~\cite{zhao2026gemgenerativesupervisionhelps} (0.200M)
      & EgoDex~\cite{hoque2026egodexlearningdexterousmanipulation} (90.000M) \\
    \bottomrule
  \end{tabularx}
\end{table}

To preserve pre-trained semantic priors, the Vision Encoder, Understanding Expert, and Action Expert remain frozen; only the Affordance Generation Expert, learnable queries, and decoders are optimized. The model is supervised by a multi-task affordance objective:
\begin{equation}
\mathcal{L}_{\text{Stage1}} = \lambda_{\text{which}}\mathcal{L}_{\text{which}} + \lambda_{\text{where}}\mathcal{L}_{\text{where}} + \lambda_{\text{shape}}\mathcal{L}_{\text{shape}} + \lambda_{\text{layout}}\mathcal{L}_{\text{layout}}
\label{eq:stage1_loss}
\end{equation}
where weights ($\lambda_{\text{which, where, shape}}=0.1$, $\lambda_{\text{layout}}=0.04$) balance optimization across physical dimensions. Although the branch-specific targets refer to the same task-conditioned object, it contains no explicit cross-dimensional consistency or reprojection term. The resulting prediction-level coherence is evaluated in \Cref{sec:supp_2d3d_consistency}.

\subsubsection{Stage II: Affordance-Augmented Robotic Data Co-Training.}

Building upon the initial alignment, Stage II performs a rigorous re-alignment of affordance generation and robotic execution by co-training the entire AffordanceVLA on Large-Scale Synthetic Robotic Data (e.g., InternData-A1~\cite{interndata}). Crucially, the perfect annotation of high-quality data maximizes the inherent power within the dataset. During this stage, the Understanding and Action Experts are unfrozen for end-to-end alignment, while the Vision Encoder is fine-tuned with a reduced learning rate. To address the scarcity of affordance labels, we engineered an automated data processing pipeline (Please refer to the \textbf{supplementary materials} for detailed content and analysis.). Specifically, we extract keyframes rule-based on action sequences~\cite{bridgevla}. A text LLM (Claude Opus~4.5) then decomposes the global instruction into per-keyframe sub-instructions, and a vision--language model (Qwen3-VL~\cite{qwen3}) converts each keyframe into a detection category and a spatial affordance query. The resulting queries guide RexOmni~\cite{rexomni}, adapted on the embodied-manipulation subset of the Stage-I Pointing pool (\Cref{tab:stage1_data_pool}); its outputs are integrated with SAM~\cite{sam} and SAM-3D~\cite{sam3d}, yielding over 100,000 affordance annotations. The joint training objective is:
\begin{equation}
    \mathcal{L}_{\text{Stage2}} = \lambda_{\text{act}} \mathcal{L}_{\text{act}} + \lambda_{\text{afd}} \mathcal{L}_{\text{afd}}
    \label{eq:stage2_loss}
\end{equation}
where $\mathcal{L}_{\text{afd}}$ aggregates the losses from Stage I. 
We set $\lambda_{\text{act}}=1.0$ and $\lambda_{\text{afd}}=0.5$, slightly down-weighting the affordance objective relative to policy execution while retaining strong intermediate foresight supervision.

During Stages~II--III, we mildly perturb the intermediate affordance features fed to the Action Expert only during policy training, while using the original predictions at inference. This strategy is evaluated in \Cref{sec:cascaded_robustness}.

\subsubsection{Stage III: Target Task Post-Training.}

The final stage adapts the generalized VLA policy to specific downstream environments, such as the LIBERO~\cite{libero} and CALVIN~\cite{calvin} benchmarks. 
Maintaining the same trainable parameters and loss formulation as Stage II (\Cref{eq:stage2_loss}), but with the affordance weight further annealed to $\lambda_{\text{afd}}=0.15$ to prioritize precise control adaptation, this targeted post-training enables the model to deeply adapt its semantic alignment and affordance reasoning to the unique visual distributions, camera viewpoints, and kinematic constraints of the target platforms. 
This ensures high-fidelity, environment-aware action generation across diverse robotic settings. For real-world deployment, this stage further includes a brief post-training phase on a subset of the DROID dataset~\cite{khazatsky2025droidlargescaleinthewildrobot} to narrow the sim-to-real gap before the final in-house fine-tuning.

\subsubsection{Design Philosophy.}
Underlying this curriculum is a data-centric view of what supervision should carry. By enriching robotic trajectories with structured affordance labels, each demonstration encodes not merely \textit{what} to do (the action) but also \textit{how} to do it---which object to engage, where to interact, and the target's geometry. Crucially, because the affordance objective is anchored to the vision--language semantics rather than to raw control, it supervises the backbone with a signal that is far more informative than the action loss alone, helping the VLM retain its instruction-following ability instead of having it eroded by control-only optimization.

\begin{findingbox}
Rich annotations should encode not just \textit{what} to do but \textit{how} to do it; structured affordance supervision preserves the backbone's vision--language ability instead of eroding it under the action loss (\cf $\pi_{0.5}$/$\pi_{0.7}$).
\end{findingbox}

\section{Experiment}
\label{sec:exp}

We conduct extensive experiments on both simulation benchmarks and real-world tasks to comprehensively evaluate our proposed AffordanceVLA. These experiments not only validate the model's performance but also aim to answer the following questions:
\begin{itemize}
    \item \textbf{Q1 (Representation Strategy):} Can structured affordance forecasting serve as a more effective intermediate representation?
    Moreover, can the Action Expert remain robust to imperfect affordance predictions rather than forming a brittle cascaded pipeline?
    \item \textbf{Q2 (Architecture Design):} Does the decoupled design of the Mixture-of-Transformer (MoT) architecture, with specialized Understanding, Affordance Generation, and Action experts, effectively prevent representation collapse and outperform unified network structures?
    \item \textbf{Q3 (Training Paradigm):} How does the proposed three-stage progressive training strategy contribute to bridging the gap between broad visual-language pre-training and task-specific embodied control?
    \item \textbf{Q4 (Scalability):} Does scaling Stage~I VQA jointly improve affordance quality and OOD robot control?
    Can scalable 2D bootstrap 3D affordance learning while preserving 2D–3D consistency without explicit constraints?
\end{itemize}
Beyond aggregate performance, \textbf{Q1} and \textbf{Q4} explicitly examine two central concerns of structured intermediate prediction: error propagation to downstream control and coherence between the separately predicted 2D and 3D affordances.
\subsection{Baseline Comparison}

\subsubsection{Simulation Setup.}
We evaluate standard control on LIBERO~\cite{libero} and CALVIN~\cite{calvin}, kitchen multitask transfer on RoboCasa~\cite{nasiriany2024robocasalargescalesimulationeveryday}, and controlled OOD evaluation on LIBERO-Plus~\cite{fei2025liberoplusindepthrobustnessanalysis}.
To better isolate the contributions of our architecture and training paradigm, we report the performance of two model variants in our main results: 

1) \textbf{AffordanceVLA (w/o stage II):} A variant that skips Stage II (affordance-augmented robotic data co-training), proceeding directly from Stage I (general affordance grounding pre-training) to Stage III (target task fine-tuning). 

2) \textbf{AffordanceVLA (full):} The complete model trained with our full three-stage progressive strategy (Stage I $\rightarrow$ Stage II $\rightarrow$ Stage III).

\subsubsection{LIBERO Benchmark.}

As shown in Table~\ref{tab:libero}, our full model attains a strong average success rate of 95.8\%---the highest among the methods compared here and competitive with the best recent VLAs. This consistent performance across all four suites validates the effectiveness and robustness of the AffordanceVLA framework.
While AffordanceVLA excels on the Spatial, Object, and Goal suites, the margin narrows on LIBERO-Long (89.8\%), suggesting that extremely long-horizon sequential tasks would further benefit from explicit long-term memory---a limitation shared across current VLAs that we revisit qualitatively in our real-world long-horizon study (\Cref{tab:real_world_combined}).

\begin{table}[tb]
  \centering
  \small
  \renewcommand{\arraystretch}{0.95}
  \setlength{\tabcolsep}{4pt}
  \caption{\textbf{Quantitative Results on the LIBERO Benchmark.} We report the success rates (\%) across four distinct task suites: Spatial, Object, Goal, Long, along with the average success rate. All methods are evaluated over 50 rollouts. AffordanceVLA outperforms baselines with a higher average success rate. The best results are \textbf{bolded}.}
  \label{tab:libero}

  \begin{adjustbox}{max width=1\linewidth}
  \begin{tabular}{@{}l cccc c@{}}
    \toprule
    Method & Spatial & Object & Goal & Long & \textbf{Average} \\
    \midrule
    OpenVLA~\cite{openvla} & 84.7 & 88.4 & 79.2 & 53.7 & 76.5 \\
    SpatialVLA~\cite{spatialvla} & 88.2 & 89.9 & 78.6 & 55.5 & 78.1 \\
    CoT-VLA~\cite{cotvla} & 87.5 & 91.6 & 87.6 & 69.0 & 83.9 \\
    ThinkAct~\cite{thinkact} & 88.3 & 91.4 & 87.1 & 70.9 & 84.4 \\
    Pi0~\cite{pi0} & 98.0 & 96.8 & 94.4 & 88.4 & 94.4 \\
    gr00t-N1~\cite{gr00tn1} & 94.4 & 97.6 & 93.0 & 90.6 & 93.9 \\
    F1-VLA~\cite{f1vla} & 98.2 & 97.8 & 95.4 & \textbf{91.3} & 95.7 \\
    \midrule
    \textbf{AffordanceVLA (w/o stage II)} & 88.5 & 91.7 & 91.3 & 73.3 & 86.2 \\
    \textbf{AffordanceVLA (full)} & \textbf{98.6} & \textbf{98.4} & \textbf{96.2} & 89.8 & \textbf{95.8} \\
    \bottomrule
  \end{tabular}
  \end{adjustbox}
\end{table}

\subsubsection{CALVIN ABC$\rightarrow$D Benchmark.}
As shown in Table~\ref{tab:calvin}, AffordanceVLA (full) achieves strong performance with an Average Length of 4.33, completing 5 consecutive tasks in 75.9\% of the rollouts---competitive among recent VLAs on this challenging zero-shot OOD protocol.
Crucially, the substantial performance jump from AffordanceVLA (w/o stage II) (3.81) to our full model (4.33) highlights the absolute necessity of the Stage II progressive training strategy. It systematically accumulates generic world knowledge, preventing overfitting to training environments (\textbf{Q3}). Furthermore, this OOD robustness is safeguarded by the MoT architecture (\textbf{Q2}). Its attention mechanism ensures that the generalized semantic representations learned during Stage I and II are strictly protected from high-frequency control policy updates in Stage III, thereby smoothly unlocking strong generalization.

\begin{table}[tb]
  \centering
  \small
  \renewcommand{\arraystretch}{1.03}
  \setlength{\tabcolsep}{4.2pt}
  \caption{\textbf{Quantitative Results on CALVIN ABC$\rightarrow$D.} Success rates (\%) are reported for completing 1 to 5 consecutive language-conditioned tasks over 1000 rollouts. Best results are \textbf{bolded}.}
  \label{tab:calvin}

  \begin{adjustbox}{max width=1\linewidth}
  \begin{tabular}{@{}l ccccc c@{}}
    \toprule
    \multirow{2}{*}{Method} 
    & \multicolumn{5}{c}{Completed Tasks (\%)} 
    & \multirow{2}{*}{\textbf{Avg. Len}} \\
    \cmidrule(lr){2-6}
    & 1/5 & 2/5 & 3/5 & 4/5 & 5/5 & \\
    \midrule
    RoboFlamingo~\cite{roboflamingo} & 82.4 & 61.9 & 46.6 & 33.1 & 23.5 & 2.48 \\
    SuSIE~\cite{susie} & 87.0 & 69.0 & 49.0 & 38.0 & 26.0 & 2.69 \\
    GR-1~\cite{gr-1} & 85.4 & 71.2 & 59.6 & 49.7 & 40.1 & 3.06 \\
    OpenVLA~\cite{openvla} & 91.3 & 77.8 & 62.0 & 52.1 & 43.5 & 3.27 \\
    CLOVER~\cite{clover} & 96.0 & 83.5 & 70.8 & 57.5 & 45.4 & 3.53 \\
    UniVLA~\cite{univla} & 95.5 & 85.8 & 75.4 & 66.9 & 56.5 & 3.80 \\
    Pi0~\cite{pi0} & 93.8 & 85.0 & 76.7 & 68.6 & 60.1 & 3.84 \\
    Seer~\cite{seer} & 94.4 & 87.2 & 79.9 & 72.2 & 64.3 & 3.98 \\
    VPP~\cite{vpp} & 95.3 & 88.2 & 80.3 & 72.9 & 64.5 & 4.01 \\
    Seer-Large~\cite{seer} & 96.3 & 91.6 & 86.1 & 80.3 & 74.0 & 4.28 \\
    \midrule
    \textbf{AffordanceVLA (w/o stage II)} 
    & 93.4 & 84.7 & 75.4 & 68.1 & 58.9 & 3.81 \\
    \textbf{AffordanceVLA (full)} 
    & \textbf{96.8} & \textbf{92.0} & \textbf{87.5} & \textbf{80.8} & \textbf{75.9} & \textbf{4.33} \\
    \bottomrule
  \end{tabular}
  \end{adjustbox}
\end{table}

\subsubsection{RoboCasa Benchmark.}

\begin{table}[tb]
  \centering
  \small
  \renewcommand{\arraystretch}{0.95}
  \setlength{\tabcolsep}{4pt}
  \caption{\textbf{Quantitative Results on the RoboCasa Benchmark.}
  We report the average success rate (SR, \%) across 24 manipulation tasks.
  Demos/task denotes the number of task-specific demonstrations used for downstream adaptation.}
  \label{tab:robocasa}

  \begin{adjustbox}{max width=1\linewidth}
  \begin{tabular}{@{}l c c@{}}
    \toprule
    Method
    & Demos/task & \textbf{Average SR} \\
    \midrule

    gr00t-N1~\cite{gr00tn1}
    & 300 & 49.6 \\

    DP-VLA~\cite{dp_vla}
    & 3{,}000 & 57.3 \\

    UWM~\cite{unifiedworldmodels}
    & 1{,}000 & 60.8 \\

    Pi0.5~\cite{pi05}
    & 300 & 62.1 \\

    Pi0~\cite{pi0}
    & 300 & 62.5 \\


    gr00t-N1.6~\cite{gr00tn1}
    & 300 & 66.2 \\

    \midrule
    \textbf{AffordanceVLA (w/o stage II)}
    & \textbf{300} & 58.6 \\

    \textbf{AffordanceVLA (full)}
    & \textbf{300} & \textbf{65.0} \\

    \bottomrule
  \end{tabular}
  \end{adjustbox}
\end{table}

RoboCasa~\cite{nasiriany2024robocasalargescalesimulationeveryday} covers 24 atomic kitchen tasks spanning pick-and-place, articulated fixtures, appliance control, and insertion.
These tasks jointly require accurate target-object grounding, contact-region localization, and geometric interaction reasoning, making the benchmark well aligned with the structured Which2Act--Where2Act--How2Act formulation of AffordanceVLA. With same downstream data budget 300 demonstrations per task, AffordanceVLA raises its Pi0 base from 62.5 to 65.0 SR and also surpasses Pi0.5 by 2.9 points (\Cref{tab:robocasa}).
These results indicate that structured affordance supervision provides a more data-efficient interface for adapting pretrained vision-language representations to fine-grained robotic interactions.

\subsubsection{OOD Robustness on LIBERO-Plus.}

\begin{table}[tb]
  \centering
  \small
  \renewcommand{\arraystretch}{1.03}
  \setlength{\tabcolsep}{4.2pt}
  \caption{\textbf{Quantitative Results on LIBERO-Plus OOD Evaluation.}
  Success rates (SR, \%) are reported over 10,030 instances.
  }
  \label{tab:libero_plus}

  \begin{adjustbox}{max width=1\linewidth}
  \begin{tabular}{@{}l ccccccc c@{}}
    \toprule
    \multirow{2}{*}{Method}
    & \multicolumn{7}{c}{OOD Category (\%)}
    & \multirow{2}{*}{\textbf{Total}} \\
    \cmidrule(lr){2-8}
    & Camera & Robot & Language & Light
    & Background & Noise & Layout
    & \\
    \midrule

    OpenVLA~\cite{openvla}
    & 0.8 & 3.5 & 23.0 & 8.1 & 34.8 & 15.2 & 28.5
    & 15.6 \\

    WorldVLA~\cite{worldvla}
    & 0.1 & 27.9 & 41.6 & 43.7 & 17.1 & 10.9 & 38.0
    & 25.0 \\

    NORA~\cite{NORA}
    & 2.2 & 37.0 & 65.1 & 45.7 & 58.6 & 12.8 & 62.1
    & 39.0 \\

    UniVLA~\cite{univla}
    & 1.8 & 46.2 & 69.6 & 69.0 & 81.0 & 21.2 & 31.9
    & 42.9 \\

    Pi0~\cite{pi0}
    & 13.1 & 6.8 & 65.6 & 87.0 & 81.8 & 16.4 & 72.7
    & 45.5 \\

    Pi0.5~\cite{pi05}
    & \textbf{67.3} & \textbf{78.8} & 86.3
    & \textbf{95.8} & \textbf{93.8} & \textbf{87.3} & 86.4
    & \textbf{84.2} \\

    \midrule
    \textbf{AffordanceVLA (w/o stage II)}
    & 16.1 & 6.2 & 55.8 & 79.3 & 81.7 & 21.0 & 72.5
    & 44.2 \\

    \textbf{AffordanceVLA (full)}
    & 35.0 & 29.0 & \textbf{86.5} & 92.7 & 93.4 & 34.5
    & \textbf{88.9}
    & 62.9 \\

    \bottomrule
  \end{tabular}
  \end{adjustbox}
\end{table}

LIBERO-Plus~\cite{fei2025liberoplusindepthrobustnessanalysis} stress-tests seven controlled distribution shifts over 10,030 instances. As shown in \Cref{tab:libero_plus}, AffordanceVLA improves its Pi0 base from 45.5\% to 62.9\% total SR.
The substantial Camera and Robot gains are primarily attributed to Stage~II co-training on the diverse InternData-A1 trajectories, whose broad camera configurations and embodiment coverage complement the corresponding weaknesses of Pi0.
The remaining categories more directly validate our structured affordance design. For Language shifts, instruction-conditioned affordances act as semantic anchors, mapping diverse linguistic expressions onto stable interaction targets rather than directly coupling surface-level language variations with control. Under Light, Background, and Noise shifts, affordance prediction suppresses task-irrelevant visual variations and preserves features more intrinsically associated with interaction and action. Layout generalization further benefits from spatially grounded, object-centric priors that explicitly capture target objects and their interaction geometry. Consistent with this interpretation, AffordanceVLA substantially improves over its w/o Stage~II variant across all five categories and, despite the uncontrolled cross-base comparison, slightly surpasses Pi0.5 on both Language (86.5 vs.\ 86.3) and Layout (88.9 vs.\ 86.4).

\subsection{Ablation Studies}
\label{sec:ablation}

To systematically isolate the contributions of our proposed components and answer the questions posed in \Cref{sec:exp}, we conduct comprehensive ablation studies on the LIBERO~\cite{libero} and CALVIN ABC$\rightarrow$D~\cite{calvin} benchmarks. As shown in \cref{tab:ablation}, we evaluate our framework from the following three perspectives:

\subsubsection{Architecture Design \& Training Strategy.}
First, we investigate the necessity of the Affordance Generation expert and our three-stage training strategy, \textbf{specifically addressing Q2 and Q3}. A natural concern is \emph{where} the improvement truly originates---the high-quality data, the added supervision density, or the structured representation itself. To disentangle these competing explanations, we design a set of controls that each hold one factor fixed; we name them explicitly for clarity.

\noindent\textbf{(i) Data-Only Control (No-Afd, Pi0 Arch).} We train a plain Pi0 architecture on the same Stage~II InternData-A1 data \emph{without} the affordance objective or the Affordance Generation expert, isolating the effect of high-quality robotic data alone. It improves only marginally over the vanilla Pi0 baseline (LIBERO $92.4\%$, CALVIN $3.93$ vs.\ Pi0's $3.84$) and remains far below the full model. \textbf{Hence the gain cannot be attributed to data volume per se}---data alone is insufficient to bridge the spatial gap.

\noindent\textbf{(ii) Frozen-Representation Control (Frozen-Afd).} Freezing the Affordance Generation expert after Stage~I---\ie treating affordance as a fixed external prior fed to the Action expert---triggers a severe collapse (LIBERO $67.1\%$, CALVIN $2.83$). This confirms that affordance must be \emph{co-optimized} with the control policy: a statically pre-trained representation cannot adapt to the embodied control space, reproducing exactly the structural mismatch we set out to resolve, and sharply distinguishing our internalized affordance from prior external-cue pipelines.

\noindent\textbf{(iii) Stage-II Ablation (w/o Stage~II).} Skipping the affordance-augmented co-training markedly hurts Out-of-Distribution (OOD) generalization (CALVIN $3.81$), evidencing the role of large-scale, affordance-centric robotic data as an inductive bridge between broad vision--language understanding and task-specific embodied control \textbf{(Q3)}.

Taken together, controls (i) and (ii) \textbf{directly answer Q2}: it is the \emph{decoupled, jointly-optimized} MoT design---not merely more data nor an off-the-shelf affordance module---that prevents representation collapse and unlocks the gains.

\begin{table}[htb]
  \renewcommand{\arraystretch}{1.0}
  \setlength{\tabcolsep}{5pt}
  \caption{\textbf{Ablation Study on LIBERO and CALVIN ABC$\rightarrow$D Benchmarks.} We report success rates (\%) and their averages on four LIBERO task suites. On CALVIN, we report the success rates for multi-step task sequences ($1/5$, $3/5$, $5/5$) and the average completed chain length (Avg.\ Len). The best results are in \textbf{bold}.}
  \label{tab:ablation}
  \centering
  \resizebox{\linewidth}{!}{
    \begin{tabular}{@{}l ccccc !{\vrule width 0.5pt} cccc@{}}
      \toprule
      & \multicolumn{5}{c}{\textbf{LIBERO} (Success Rate \%)} & \multicolumn{4}{c}{\textbf{CALVIN ABC$\rightarrow$D}} \\
      \cmidrule(lr){2-6} \cmidrule(lr){7-10}
      Method & Spatial & Object & Goal & Long & Avg. & 1/5 & 3/5 & 5/5 & Avg.\ Len \\
      \midrule
      
      \rowcolor{pale_pink} 
      \multicolumn{10}{l}{\textit{Architecture Design \& Training Strategy}} \\
      No-Afd (Pi0 Arch)  & 96.0 & 95.4 & 92.4 & 85.8 & 92.4 & 94.5 & 78.0 & 62.8 & 3.93 \\
      Frozen-Afd         & 68.0 & 71.1 & 66.4 & 62.9 & 67.1 & 85.3 & 55.9 & 26.3 & 2.83 \\
      AffordanceVLA w/o stage II    & 88.5 & 91.7 & 91.3 & 73.3 & 86.2 & 93.4 & 75.4 & 58.9 & 3.81 \\
      \midrule
      
      \rowcolor{pale_pink} 
      \multicolumn{10}{l}{\textit{Affordance Representation}} \\
      w/o Which2Act & 97.5 & 97.6 & 95.0 & 88.1 & 94.6 & 96.7 & 83.3 & 72.1 & 4.20 \\
      w/o Where2Act & 95.5 & 96.0 & 93.4 & 88.0 & 93.2 & 96.2 & 81.9 & 69.8 & 4.13 \\
      w/o How2Act   & 96.1 & 96.5 & 93.9 & 88.2 & 93.7 & 95.0 & 79.4 & 65.9 & 4.01 \\
      \midrule
      
      \rowcolor{pale_pink} 
      \multicolumn{10}{l}{\textit{Attention Mechanism for Affordance}} \\
      Block-wise Tokens  & 92.4 & 92.9 & 89.8 & 86.0 & 90.3 & 94.1 & 77.1 & 61.7 & 3.89 \\
      \midrule
      
      \textbf{AffordanceVLA (full)} & \textbf{98.6} & \textbf{98.4} & \textbf{96.2} & \textbf{89.8} & \textbf{95.8} & \textbf{96.8} & \textbf{87.5} & \textbf{75.9} & \textbf{4.33} \\
      \bottomrule
    \end{tabular}
  }
\end{table}

\subsubsection{Affordance Representation.}
Next, we ablate the internal affordance representation to answer \textbf{Q1}. The observed drops align with the role of each component: Which2Act predicts task-conditioned, object-centric visual latents to suppress distractors (CALVIN $4.33\!\rightarrow\!4.20$), Where2Act estimates a 2D interaction map for precise contact localization (LIBERO $95.8\%\!\rightarrow\!93.2\%$), and How2Act models object shape and 3D spatial layout for geometric 6-DoF reasoning (CALVIN $4.33\!\rightarrow\!4.01$).
Importantly, removing any \emph{single} head yields only a \emph{graceful} degradation rather than a catastrophic collapse.
Importantly, removing any single component causes graceful rather than catastrophic degradation, indicating that the three cues are jointly consumed rather than connected by a hard deterministic chain. This design limits, but does not eliminate, the possibility that imperfect affordance predictions bias downstream actions, which we examine in \Cref{sec:cascaded_robustness}.
We further observe that How2Act's benefit is comparatively modest under the simple tabletop, two-finger settings of LIBERO/CALVIN, yet becomes pronounced on complex real-world 6-DoF interactions (\Cref{tab:real_world_combined}), precisely where 3D shape and layout priors are most valuable.
The progressive synergy of these three components---integrating information from coarse to fine granularity and from 2D to 3D representations---is key to our strongest results, \textbf{firmly establishing that structured affordance forecasting serves as a more effective intermediate representation (Q1).}
Detailed analysis of the effectiveness of affordance subgoals can be found in the \textbf{supplementary material}.

\subsubsection{Robustness to Cascaded Affordance Errors.}
\label{sec:cascaded_robustness}

\begin{wrapfigure}[31]{r}{0.435\textwidth}
  \vspace{-0.8\baselineskip}
  \centering

  \captionof{table}{\small Training Configurations of Methods on Data Efficiency Analysis.}
  \label{tab:experimental_settings}
  \renewcommand{\arraystretch}{1.0}
  \setlength{\tabcolsep}{3.2pt}

  \resizebox{\linewidth}{!}{
    \begin{tabular}{lccc}
      \toprule
      \textbf{Method} & \textbf{Stage I} & \textbf{Stage II} & \textbf{Stage III} \\
      \midrule
      \textcolor{colorPi0}{\textbf{Pi0}} & \scalebox{0.8}{$\times$} & \scalebox{0.8}{$\times$} & Phased \\
      \textcolor{colorPi0}{\textbf{(Finetuned)}} & & & (5k$\sim$Full) \\
      \addlinespace[0.45em]
      \textcolor{colorNoAfd}{\textbf{No-Afd}} & \scalebox{0.8}{$\times$} & Action loss & Phased \\
      \textcolor{colorNoAfd}{\textbf{(Pi0 Arch)}} & & Only & (5k$\sim$Full) \\
      \addlinespace[0.45em]
      \textcolor{colorOursScratch}{\textbf{AffordanceVLA}} & VQA & \scalebox{0.8}{$\times$} & Phased \\
      \textcolor{colorOursScratch}{\textbf{(w/o stage II)}} & Afford. & & (5k$\sim$Full) \\
      \addlinespace[0.45em]
      \textcolor{colorOursPretrained}{\textbf{AffordanceVLA}} & VQA & Act loss + & Phased \\
      \textcolor{colorOursPretrained}{\textbf{(full)}} & Afford. & Afd loss & (5k$\sim$Full) \\
      \bottomrule
    \end{tabular}
  }
  \vspace{4mm}

  \includegraphics[width=\linewidth]{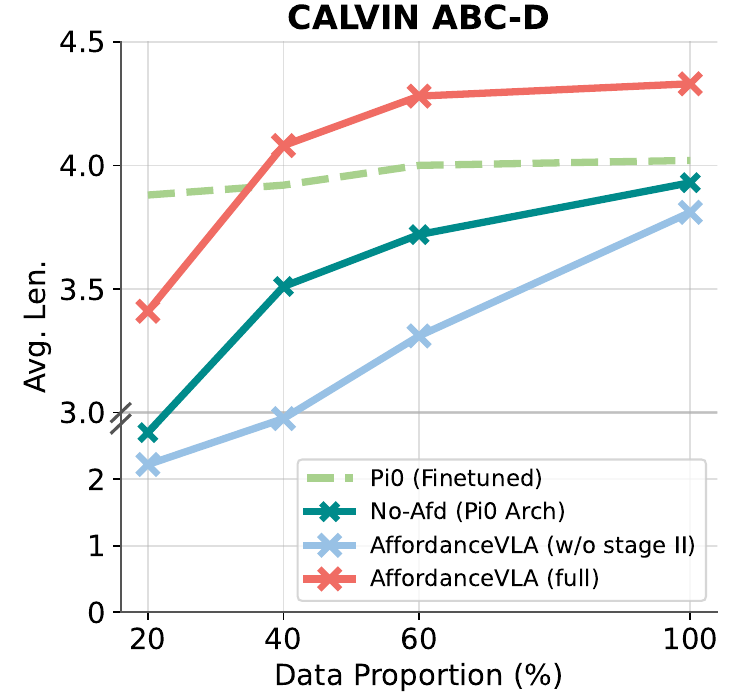}
  \vspace{-5mm}
   \captionsetup{
    type=figure,
    justification=centering,
    singlelinecheck=false
  }
  \caption{\textbf{Data Efficiency Curve.}}
  \label{fig:data_efficiency}
  
\end{wrapfigure}

A residual risk of intermediate-prediction frameworks is that upstream errors may bias downstream action generation~\cite{yang2026trustworthyembodiedintelligencesystems}. To improve robustness, we stochastically perturb the intermediate affordance features supplied to the Action Expert during policy training, while applying no additional perturbation at inference. This improves the LIBERO-Plus total OOD SR from 61.3\% without perturbation to 62.9\% with perturbation (+1.6 points), indicating that exposure to imperfect intermediate cues reduces the Action Expert's sensitivity to upstream prediction errors. Rather than explicitly correcting erroneous affordance predictions, this regularization enables downstream control to absorb small deviations instead of amplifying them.

\subsubsection{Attention Mechanism for Affordance (Same-Density Control).}
As mentioned in \Cref{attention_mechanism}, bidirectional intra-expert attention ensures thorough contextual fusion within the Affordance Generation expert. 
This mechanism also enables our most pointed control for the \emph{representation vs.\ supervision-density} question: the \textbf{Block-wise Tokens} variant keeps the affordance losses and the data---and therefore the total \emph{supervision density}---exactly identical, and only replaces the bidirectional attention with a causal scheme so that the three modules (Which2Act, Where2Act, and How2Act) are strictly prohibited from cross-attending to each other. This degenerates the organic affordance representation into three independent, equally-dense auxiliary tasks.
As shown in \Cref{tab:ablation}, this single change causes a sharp decline to 90.3\% on LIBERO and 3.89 on CALVIN. \textbf{Because the supervision density is held constant, this directly rules out the ``multi-task density'' explanation}: what drives the gain is the structured, jointly-refined representation, not the mere addition of more loss terms.
Importantly, the block-wise variant uses exactly the same training data, affordance annotations, and prediction objectives as the full model, differing only in whether tokens across affordance dimensions can interact. Its degradation therefore shows that the gain cannot be attributed merely to additional annotated data; cross-dimensional information exchange is itself critical for integrating these supervisory signals into a unified perception--prediction representation. Although this ablation alone does not establish a causal direction, together with the component-wise scaling dynamics in \Cref{sec:scaling_affordance}, it supports the interpretation that stable object-centric and image-plane anchors facilitate the more difficult 3D affordance learning.

\subsubsection{Scaling Affordance Learning from VQA Supervision to Robot Control.}
\label{sec:scaling_affordance}

\begin{figure*}[t]
    \centering
    \includegraphics[width=\textwidth]{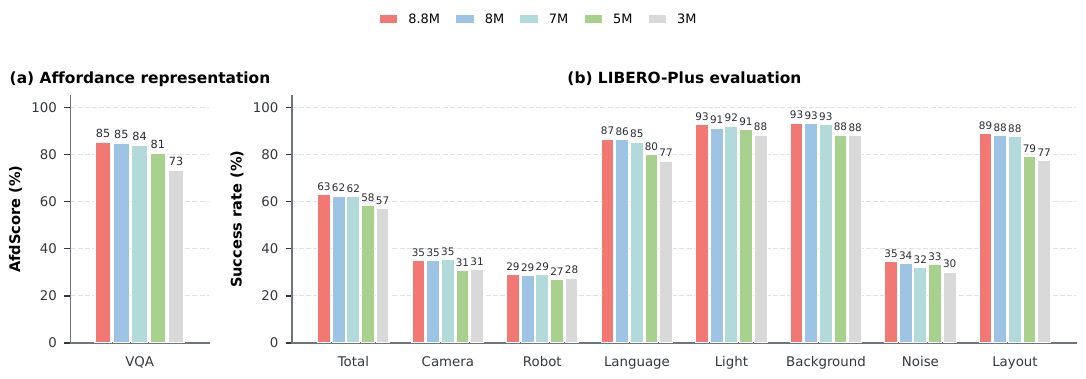}
    \caption{\textbf{Scaling VQA Supervision.} Held-out AfdScore and LIBERO-Plus OOD success versus Stage~I data scale. Metric and split details are in \Cref{sec:afdscore}.}
    \label{fig:scaling_affordance}
\end{figure*}

Scaling Stage~I VQA from 3M to 8.8M samples increases AfdScore from approximately 73 to 85 and raises the LIBERO-Plus total OOD SR from approximately 57\% to 63\% (\Cref{fig:scaling_affordance}). The clearest OOD gains occur under Language ($77\!\rightarrow\!87$), Light ($88\!\rightarrow\!93$), Background ($88\!\rightarrow\!93$), and Layout ($77\!\rightarrow\!89$) shifts. Camera, Robot, and Noise vary less or non-monotonically, suggesting that their absolute robustness is influenced more strongly by the fixed robotic co-training data than by additional Stage~I VQA supervision. Overall, these results establish the primary scaling chain from scalable VQA supervision to stronger affordance representations and, ultimately, more robust robot control.

\noindent\textbf{2D-to-3D Scaling and Coherence.}
Actually, the component-wise AfdScore internal data reveal a clear scaling asymmetry: Which2Act and Where2Act improve steadily, while How2Act Shape and Layout dominate the early bottleneck and training fluctuations. This indicates that 3D geometry and spatial-layout reasoning are harder to scale than object-centric grounding and 2D interaction localization. At larger data scales, however, both 3D components stabilize and reach their best performance, suggesting a learnable data-and-optimization bottleneck rather than a fixed representational ceiling. Since our pipeline lifts scalable 2D localization into aligned SAM-3D targets, these trends support a 2D-to-3D bootstrapping interpretation. Moreover, \Cref{sec:supp_2d3d_consistency} shows that the resulting 2D and 3D predictions remain coherent even without an explicit consistency objective. Together, these results answer \textbf{Q4}.

\begin{findingbox}
\textbf{Finding.} Scalable VQA affordance supervision $\rightarrow$ stronger affordance representations $\rightarrow$ Scalable robot control.
\end{findingbox}

\subsubsection{Data Efficiency.}

We evaluate data efficiency by scaling the downstream fine-tuning data from 10\% to 100\%; the four training configurations under comparison are summarized in \Cref{tab:experimental_settings}, and the corresponding curves are reported in \Cref{fig:data_efficiency}.

\noindent\textbf{Initial distribution shift.} At the extreme low-data regime (20\%), the vanilla \textbf{Pi0} starts strong thanks to its original pre-trained weights, whereas all models that undergo specialized pre-training---\textbf{No-Afd}, \textbf{AffordanceVLA (w/o stage II)}, and the full \textbf{AffordanceVLA}---begin lower. This dip is an expected consequence of the temporary weight misalignment introduced by their respective pre-training stages.

\noindent\textbf{Why AffordanceVLA surges.}
Despite this initial shift, the full model exhibits a rapid surge and superior sample efficiency: with only 40\% of the fine-tuning data, it already attains above 4.0 on CALVIN, shattering the ceiling of the fully fine-tuned \textbf{Pi0}.
We attribute this to the structured affordance representation (Which2Act $+$ Where2Act $+$ How2Act), which decomposes the perception--action mapping into interpretable sub-problems, so that each additional sample supervises not only the final action but also object grounding, spatial localization, and 3D geometric reasoning---effectively multiplying the learning signal per sample.

\noindent\textbf{Ablation trajectories.} The recovery trajectories of the ablations are equally telling. \textbf{No-Afd} recovers slowly and struggles to clearly surpass the original \textbf{Pi0}, indicating that robotic pre-training data alone offers limited downstream scalability without affordance structure; \textbf{AffordanceVLA (w/o stage II)} suffers the most severe shift and the slowest recovery, reaffirming Stage~II as an indispensable bridge between broad vision--language alignment and sample-efficient embodied adaptation.

\noindent\textbf{The takeaway.} Rather than reading this as a deficiency of any baseline, we read it as a statement about \emph{representation}: the datasets themselves are not the bottleneck---the representation is. Architectural innovation that structurally grounds manipulation through affordances is what converts additional data into proportional gains, so that architecture, representation, and data quality become mutually amplifying rather than mutually substitutable.

\begin{findingbox}
Architecture unlocks data potential: Pi0 saturates while we break its ceiling at 40\% data---architecture $\times$ representation $\times$ data are mutually amplifying.
\end{findingbox}

\subsection{Real-world Experiments}

\subsubsection{Basic Task Performance.}
To bridge the sim-to-real gap, our policies are initially pre-trained on a subset of DROID dataset \cite{khazatsky2025droidlargescaleinthewildrobot} and detailed experimental setups are provided in the \textbf{supplementary material}. As shown in Table \ref{tab:real_world_combined}, AffordanceVLA exhibits strong generalizability in \textit{Basic Tasks}, achieving an average success rate of 88.3\% and consistently outperforming the Pi0 \cite{pi0} baseline across diverse objects, spatial relations, and semantic categories (e.g., color, shape).

\subsubsection{Instruction Sensitivity via Affordance Grounding.}
\begin{figure}[!t]
    \centering
    \includegraphics[width=\linewidth]{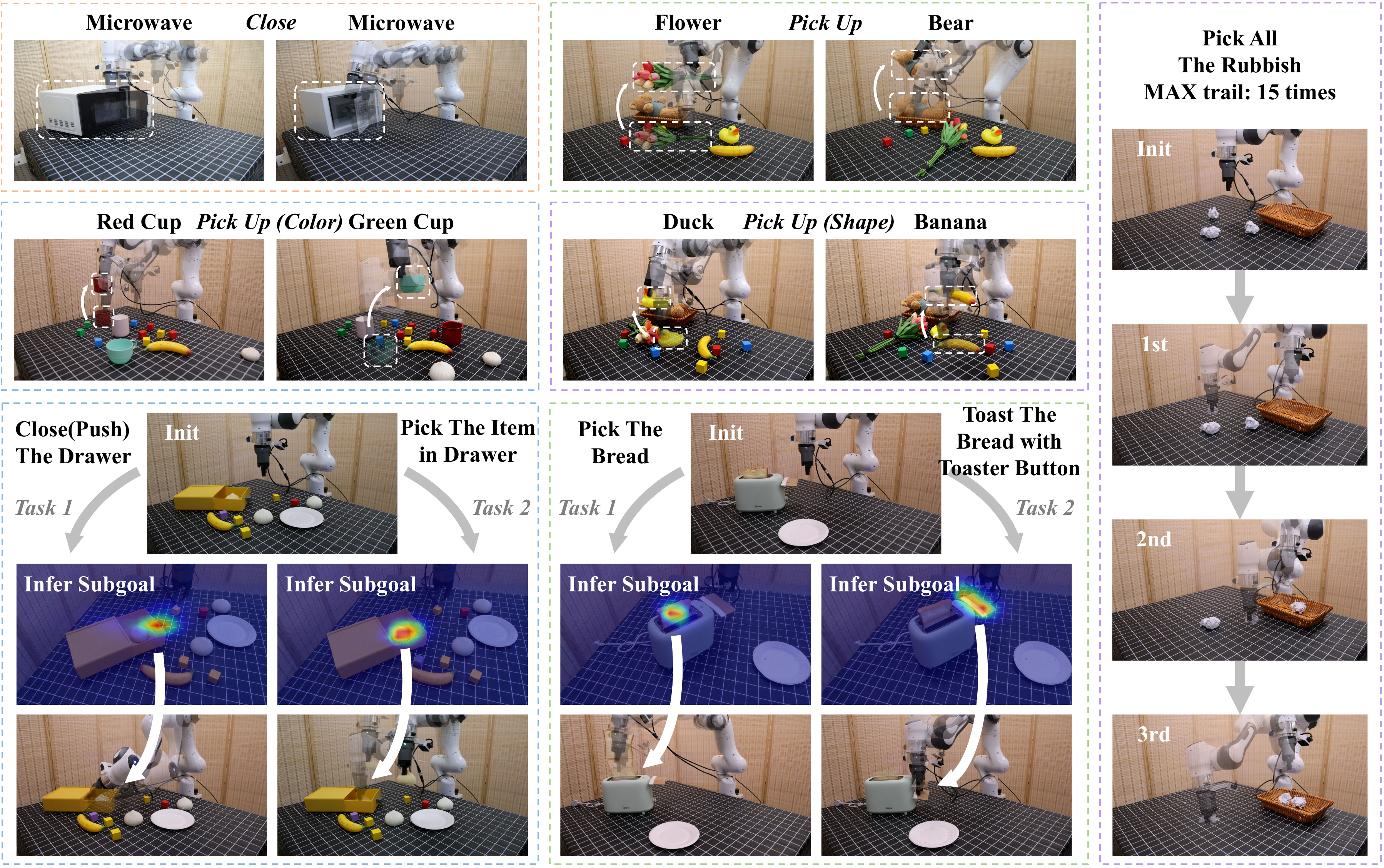}
    \caption{\textbf{Real-World Experiment Visualizations.} \textbf{Top:} Qualitative results for Basic tasks. \textbf{Bottom Left:} Visualizations of Where2Act token in the \textit{Drawer} and \textit{Toaster} tasks. \textbf{Right:} Sequential execution of the continuous \textit{Pick all the rubbish} task.}
\label{fig:real_world_1}
\end{figure}

To verify the sensitivity of our V--L bridge under severe visual aliasing, we evaluate \textit{Complex Tasks} like \textit{Drawer} and \textit{Toaster} (Fig.~\ref{fig:real_world_1}). Given identical initial visual observations but distinct instructions (\eg ``pick'' vs.\ ``close''), AffordanceVLA decisively outperforms Pi0. For instance, our model achieves 86.7\% and 100.0\% success rates in the \textit{Drawer (pick)} and \textit{Drawer (close)} tasks, compared to Pi0's 46.7\% and 40.0\%, and reaches an aggregate complex-task average of 82.9\% versus Pi0's 44.8\%. Crucially, because the two policies share identical visual inputs and differ \emph{only} in language, this gap isolates instruction-following fidelity from visual perception. It thus provides concrete empirical support for our hypothesis (discussed below) that affordance supervision preserves the VLM's vision--language ability rather than letting it be eroded by the action loss: by unambiguously grounding concise linguistic intents into highly localized affordance heatmaps, our model reliably disambiguates the required action.

\begin{table}[htbp]
  \centering
  \renewcommand{\arraystretch}{1.0}
  \setlength{\tabcolsep}{5pt}
  \caption{\textbf{Results on Real-world Tasks.} We report the average success ratio (\%) over 15 trials per task. Tasks are categorized into Basic and Complex scenarios. The best results are denoted in \textbf{bold}.}
  \label{tab:real_world_combined}
  \resizebox{\linewidth}{!}{
  \begin{tabular}{@{}l cc cc cc cc c@{}}
    \toprule
    \textbf{Method} & \multicolumn{8}{c}{\textbf{Task-specific Success Rates}} & \textbf{Average} \\
    \midrule
    
    \rowcolor{pale_pink} 
    \multicolumn{10}{l}{\textit{Basic Real-world Tasks}} \\ 
    & \multicolumn{2}{c}{Close} & \multicolumn{2}{c}{Pick up (Color)} & \multicolumn{2}{c}{Pick up (Shape)} & \multicolumn{2}{c}{Pick up} & \\
    \cmidrule(lr){2-3} \cmidrule(lr){4-5} \cmidrule(lr){6-7} \cmidrule(lr){8-9}
    & microwave & safe & red & green & duck & banana & flower & bear & \\
    \midrule
    Pi0~\cite{pi0} & 86.7 & 86.7 & 80.0 & 80.0 & 26.7 & 73.3 & 53.3 & 80.0 & 70.8 \\
    \textbf{AffordanceVLA} & \textbf{93.3} & \textbf{100.0} & \textbf{86.7} & \textbf{80.0} & \textbf{86.7} & \textbf{86.7} & \textbf{80.0} & \textbf{93.3} & \textbf{88.3} \\
    
    \midrule
    \addlinespace[0.5em]
    
    \rowcolor{pale_pink}
    \multicolumn{10}{l}{\textit{Complex Real-world Tasks}} \\
    & \multicolumn{2}{c}{Drawer} & \multicolumn{2}{c}{Toaster} & \multicolumn{4}{c}{Pick all the rubbish} & \\
    \cmidrule(lr){2-3} \cmidrule(lr){4-5} \cmidrule(lr){6-9}
    & pick & close & pick & toast & 1st ($\uparrow$) & 2nd ($\uparrow$) & 3rd ($\uparrow$) & Empty ($\downarrow$) & \\
    \midrule
    Pi0~\cite{pi0} & 46.7 & 40.0 & 46.7 & 26.7 & 93.3 & 53.3 & 6.7 & 33 & 44.8 \\
    \textbf{AffordanceVLA} & \textbf{86.7} & \textbf{100.0} & \textbf{80.0} & \textbf{86.7} & \textbf{100.0} & \textbf{80.0} & \textbf{46.7} & \textbf{11} & \textbf{82.9} \\
    \bottomrule
  \end{tabular}
  }
\end{table}

\subsubsection{Closed-Loop Multi-Step Execution.}
The continuous \textit{Pick all the rubbish} task demonstrates how Affordance Subgoals compensate for the lack of explicit long-term planning architectures. By dynamically re-evaluating the workspace, our single-frame model generates sequential visual targets to translate broad instructions into continuous execution. While Pi0's performance sharply degrades by the 3rd execution (6.7\%), AffordanceVLA maintains a robust 46.7\% success rate. Furthermore, our affordance-driven approach drastically improves execution efficiency, reducing total redundant actions (Empty Picks) to 11 versus Pi0's 33. This intent enrichment empowers the model to naturally exhibit long-horizon capabilities.

\subsubsection{Why Does Affordance Help? A Discussion.}
\label{sec:why_affordance}
A closer look at the failure modes is revealing. In the \textit{Toaster} task, Pi0 performs poorly (toast $26.7\%$ vs.\ our $86.7\%$), and its bad cases concentrate almost entirely at the button-pressing step: instead of extending to press the button, it frequently \emph{closes the gripper} as if still executing a pick-and-place, largely disregarding the ``press the button'' instruction. This exposes that Pi0, even after real-trajectory fine-tuning, still suffers from weak instruction following---its behavior is driven by the dominant action prior rather than by the language command.

We offer an intuition---admittedly a hypothesis rather than a proven mechanism---for why affordance grounding alleviates this. In an action-only VLA such as Pi0, the low-level action loss is back-propagated directly into the VLM backbone; this signal is arguably neither sufficiently informative nor well-aligned with the VLM's vision--language semantics, and may gradually erode the instruction-following ability that large-scale pre-training endowed. Recent strong generalist policies appear to mitigate this implicitly: $\pi_{0.5}$~\cite{pi05} and $\pi_{0.7}$~\cite{pi07} introduce structured intermediate supervision (\eg discrete action tokens or bounding boxes) that is used only during training and \emph{not} decoded at deployment, plausibly acting as a semantic anchor that keeps the backbone from drifting under the control objective. We conjecture that affordance plays a similar---and arguably more natural---role: as an intrinsic vision--language--action bridge, its training signal stays close to the VLM's semantic space, anchoring the backbone while still directly serving action. This is consistent with our \textit{Drawer} and \textit{Toaster} results, where AffordanceVLA responds correctly to the language command under identical visual observations while Pi0 does not. We stress that this anchoring account is an interpretive hypothesis meant to guide intuition, not a claim backed by direct mechanistic evidence.

\begin{findingbox}
\textbf{Hypothesis.} We conjecture that affordance acts as a structured semantic anchor: rather than letting the low-level action loss reshape the VLM directly, the affordance objective---being close to vision--language semantics---helps preserve the backbone's instruction-following ability, in spirit with the train-only intermediate cues adopted by recent strong VLAs.
\end{findingbox}

\section{Conclusion}
\label{sec:conclusion}

In this work, we present AffordanceVLA to bridge the structural gap between the semantic space of Vision-Language Models and the 3D physical requirements of embodied control. Instead of relying on direct end-to-end mappings or redundant visual foresight, our framework adopts affordances as a task-oriented intermediate representation and decomposes affordance forecasting into Which2Act, Where2Act, and How2Act. With a Mixture-of-Transformer architecture and a progressive data curriculum, AffordanceVLA achieves strong, competitive performance on ID \& OOD simulation and real-world experiments, demonstrating strong generalization and robust reasoning, validating structured affordance as an effective and scalable intermediate representation for robot control.. Future work will explore explicit temporal modeling as well as extensions to bimanual and deformable object manipulation.

\FloatBarrier
\bibliographystyle{assets/plainnat}
\bibliography{main}

\clearpage
\beginappendix
\paragraph{\textbf{Outline}} 
In this supplementary material, we provide additional details and extra experimental results to support the main manuscript:

\begin{enumerate}
    \item \textbf{\Cref{sec:afdscore}: AfdScore Evaluation Protocol}
    \item \textbf{\Cref{sec:supp_2d3d_consistency}: Evaluation-Only 2D--3D Consistency}
    \item \textbf{\Cref{sec:eff_aff}: The Effectiveness of Affordance Subgoals}
    \begin{itemize}
        \item \textbf{\Cref{sec:subgoals_quantitative}}: Prev-Variant Quantitative Diagnostics
        \item \textbf{\Cref{sec:subgoals_decoupling}}: Representation Decoupling: Backbone vs. Decoder
        \item \textbf{\Cref{sec:subgoals_qualitative}}: Qualitative Analysis of Affordance Grounding
    \end{itemize}


    \item \textbf{\Cref{sec:supp_arch}: Model Variants and Design Choice Details}
    \begin{itemize}
        \item \textbf{\Cref{sec:design_insight}}: Design Insight
        \item \textbf{\Cref{sec:which2act_redesign}}: Which2Act Redesign
        \item \textbf{\Cref{sec:model_variants}}: Model Variants
    \end{itemize}

    \item \textbf{\Cref{sec:supp_data}: Data-Centric Methodology}
    \begin{itemize}
        \item \textbf{\Cref{sec:supp_data_quality}}: Data Quality as the Performance Ceiling
        \textbf{\item \Cref{sec:supp_pipeline}: Affordance Annotation Pipeline}
    \end{itemize}

    \item \textbf{\Cref{sec:data_statistics}: Dataset Details}

    \item \textbf{\Cref{sec:training_recipe}: Training Details}

    \item \textbf{\Cref{sec:latency_breakdown}: Inference Latency}

    \item \textbf{\Cref{sec:real}: Real-World Experiments Details and Extra Experiments}

\end{enumerate}

\section{AfdScore Evaluation Protocol}
\label{sec:afdscore}

AfdScore combines four component metrics on a common $[0,100]$ scale:
\begin{itemize}
    \item \textbf{Which2Act: Latent Retrieval R@1.} Each sample has 32 fixed candidates: one target and 31 hard negatives. R@1 tests whether cosine similarity ranks the target first.
    \item \textbf{Where2Act: Object-normalized PCK@0.05.} The heatmap peak is mapped to the input image and is correct within $5\%$ of the target-box diagonal from the nearest GT contact.
    \item \textbf{How2Act--Shape: Normalized Diffusion-Denoising Score.} Using fixed evaluation timesteps and noise samples, we compute the mean noise-prediction error $\mathcal{E}_{\mathrm{den}}$ in \Cref{eq:shape} and report
    $S_{\mathrm{Shape}}=100\,\operatorname{clip}_{[0,1]}(1-\mathcal{E}_{\mathrm{den}}/\mathcal{E}_{\mathrm{null}})$,
    where $\mathcal{E}_{\mathrm{null}}=\mathbb{E}\lVert\epsilon\rVert^2$ is the no-signal denoising error under the same samples. This anchors no-signal and perfect denoising at $0$ and $100$, respectively, matching the direction and bounded percentage scale of the other components.
    We treat this score as a training-objective-aligned proxy metric for Shape quality, intended for controlled comparisons across models and data scales under the same diffusion configuration rather than as a general-purpose 3D reconstruction metric.
    The same timesteps, noise samples, and normalization are used for every model and data scale.
    \item \textbf{How2Act--Layout: 3D Layout Success Rate.} After symmetry correction, success requires $e_R\leq15^\circ$, $e_t\leq5\,\mathrm{cm}$, and $e_s\leq10\%$.
\end{itemize}
\begin{equation}
S_{\mathrm{How}}=\frac{S_{\mathrm{Shape}}+S_{\mathrm{Layout}}}{2},\qquad
\mathrm{AfdScore}=\frac{S_{\mathrm{Which}}+S_{\mathrm{Where}}+S_{\mathrm{How}}}{3}.
\label{eq:afdscore}
\end{equation}
Thus, Which and Where each contribute $1/3$, while Shape and Layout each contribute $1/6$.

\paragraph{AfdEval split.}
AfdEval is isolated from the cleaned, deduplicated VQA pool before constructing the five nested training sets:
\begin{equation}
\mathcal{D}_{3\mathrm{M}}\subset\mathcal{D}_{5\mathrm{M}}\subset
\mathcal{D}_{7\mathrm{M}}\subset\mathcal{D}_{8\mathrm{M}}\subset
\mathcal{D}_{8.8\mathrm{M}},\qquad
\mathcal{D}_{\mathrm{AfdEval}}\cap\mathcal{D}_{8.8\mathrm{M}}=\varnothing.
\end{equation}
The same AfdEval split is used at every scale. Split assignment follows source groups, never individual VQAs: all crops, prompt rewrites, and augmentations of one image; all frames of one video or clip; all frames and camera views of one robot trajectory or episode; all samples from one simulation scene seed; and all rendered views of one 3D asset remain within one split.

\FloatBarrier

\section{Evaluation-Only 2D--3D Consistency}
\label{sec:supp_2d3d_consistency}

The 2D and 3D affordance targets refer to the same task-conditioned object, but their predictions are optimized through branch-specific objectives without an explicit cross-dimensional consistency or reprojection loss. We therefore evaluate whether the learned predictions nevertheless form a coherent spatial description.

\paragraph{Evaluation protocol.}
We evaluate the five Stage-I checkpoints used in the main scaling study on a fixed set of 5,000 held-out simulator keyframes. Each sample provides camera calibration, instance masks, target identity, contact annotations, object geometry, pose, and scale. We retain only scenes containing at least two visible manipulable instances, preventing consistency from being trivially satisfied in single-object scenes. All simulator metadata are used exclusively for evaluation and are never exposed to the model or used for checkpoint selection. The same samples, decoding seeds, association rules, and thresholds are used for every checkpoint. We report scene-level bootstrap 95\% confidence intervals.

\paragraph{Cross-dimensional association.}
Let $H_n$ denote the predicted Where2Act heatmap and
\begin{equation}
    \hat{\mathbf p}_n=\arg\max_{\mathbf p}H_n(\mathbf p)
\end{equation}
its peak. Using the simulator instance masks, we associate
$\hat{\mathbf p}_n$ with a visible instance $\hat{o}^{2D}_n$; the prediction is marked invalid if no instance lies within $0.02$ times the image diagonal.

We decode the predicted Shape into its canonical 3D support and transform it using the predicted Layout:
\begin{equation}
    \hat{\mathcal V}_n =
    \hat{\mathcal T}_n(\hat{\mathcal S}_n).
\end{equation}
The corresponding 3D instance is independently determined as
\begin{equation}
    \hat{o}^{3D}_n =
    \arg\max_{o\in\mathcal O_n}
    \operatorname{IoU}_{3D}
    \left(\hat{\mathcal V}_n,\mathcal V_{n,o}\right),
\end{equation}
and is marked invalid when the maximum 3D IoU is below $0.25$. We additionally render $\hat{\mathcal V}_n$ with the calibrated camera and scene z-buffer to obtain its visible image-plane support $\hat{\mathcal M}^{3D}_n$.

\paragraph{Metrics.}
We define \emph{Cross-Dimensional Agreement} (CDA) as
\begin{equation}
\mathrm{CDA}
=
\frac{100}{N}\sum_{n=1}^{N}
\mathbf{1}
\left[
  \hat{o}^{2D}_n=\hat{o}^{3D}_n\neq\varnothing
\right]
\mathbf{1}
\left[
  \frac{
  d(\hat{\mathbf p}_n,\hat{\mathcal M}^{3D}_n)
  }{D_n}
  \leq 0.02
\right],
\end{equation}
where $D_n$ is the image diagonal. CDA therefore requires both instance-level identity agreement and geometric compatibility between the 2D interaction point and the projected 3D prediction.

CDA alone could credit two branches that consistently select the same incorrect object. We therefore introduce the stricter \emph{Target-Grounded Joint Correctness} (TGJC):
\begin{align}
\mathrm{TGJC}
=
\frac{100}{N}\sum_{n=1}^{N}
&\mathbf{1}[\mathrm{CDA}_n]
\mathbf{1}[\hat{o}^{2D}_n=o_n^\star]
\mathbf{1}
\left[
  \frac{\|\hat{\mathbf p}_n-\mathbf p_n^\star\|_2}
  {D_n^{\mathrm{box}}}
  \leq 0.05
\right] \nonumber\\
&\cdot
\mathbf{1}
\left[
  \operatorname{IoU}_{3D}
  (\hat{\mathcal S}_n,\mathcal S_n^\star)
  \geq 0.50
\right]
\mathbf{1}
\left[
  e_R\leq15^\circ,\,
  e_t\leq0.05\mathrm{m},\,
  e_s\leq10\%
\right],
\end{align}
where $D_n^{\mathrm{box}}$ is the target-box diagonal. For rotationally symmetric objects, $e_R$ is minimized over the simulator-provided symmetry group. TGJC counts a sample only when the 2D and 3D predictions agree with each other and are jointly correct with respect to the target.

As a null control, we construct a cross-branch shuffle by permuting How2Act predictions within bins matched by task category, camera, and number of visible instances. This preserves the marginal prediction distributions while destroying sample-level 2D--3D correspondence.

\begin{table}[t]
  \centering
  \small
  \setlength{\tabcolsep}{6pt}
  \renewcommand{\arraystretch}{1.05}
  \caption{\textbf{Evaluation-only 2D--3D consistency.}
  CDA measures instance- and projection-level agreement, while TGJC additionally requires both branches to be correct with respect to the ground-truth target. Values are percentages with scene-level bootstrap 95\% confidence intervals.}
  \label{tab:supp_2d3d_consistency}
  \begin{tabular}{lcc}
    \toprule
    Stage-I supervision & CDA $\uparrow$ & TGJC $\uparrow$ \\
    \midrule
    Cross-branch shuffle
      & 17.6 {\scriptsize [16.6, 18.6]}
      & 0.6 {\scriptsize [0.4, 0.9]} \\
    \midrule
    3M
      & 64.9 {\scriptsize [63.6, 66.2]}
      & 36.8 {\scriptsize [35.4, 38.2]} \\
    5M
      & 70.7 {\scriptsize [69.4, 71.9]}
      & 42.6 {\scriptsize [41.2, 44.0]} \\
    7M
      & 76.8 {\scriptsize [75.6, 78.0]}
      & 49.8 {\scriptsize [48.4, 51.2]} \\
    8M
      & 79.4 {\scriptsize [78.3, 80.5]}
      & 52.4 {\scriptsize [51.0, 53.8]} \\
    8.8M
      & \textbf{83.6} {\scriptsize [82.6, 84.6]}
      & \textbf{58.1} {\scriptsize [56.7, 59.5]} \\
    \bottomrule
  \end{tabular}
\end{table}

CDA increases from 64.9\% to 83.6\% as Stage-I supervision scales from 3M to 8.8M samples, indicating progressively stronger instance-level and geometric agreement across the 2D and 3D predictions. TGJC concurrently improves from 36.8\% to 58.1\%. Because TGJC requires both branches to match the ground-truth target under their native correctness criteria, this gain cannot be explained by the branches merely agreeing on the same incorrect object. The large gap from the cross-branch shuffle further confirms that the measured coherence is sample-specific rather than a consequence of marginal object or layout biases. These results demonstrate that prediction-level 2D--3D coherence progressively emerges under joint task-conditioned learning despite the absence of an explicit cross-dimensional consistency objective.

\section{The Effectiveness of Affordance Subgoals}
\label{sec:eff_aff}

This section reports Prev-variant quantitative diagnostics, representation-decoupling experiments, and qualitative grounding results.

\subsection{Prev-Variant Quantitative Diagnostics}
\label{sec:subgoals_quantitative}

The figures below evaluate only the Prev (VQ-VAE) variant (\Cref{tab:model_variants}): Token Accuracy for Which2Act and How2Act Shape, AUC-ROC for Where2Act, and rotation, translation, and scale errors for How2Act Layout. These metrics apply only to Prev and are not used by the current model or scaling study.

\begin{figure}[t]
    \centering
    \begin{subfigure}{\linewidth}
        \centering
        \includegraphics[width=\linewidth]{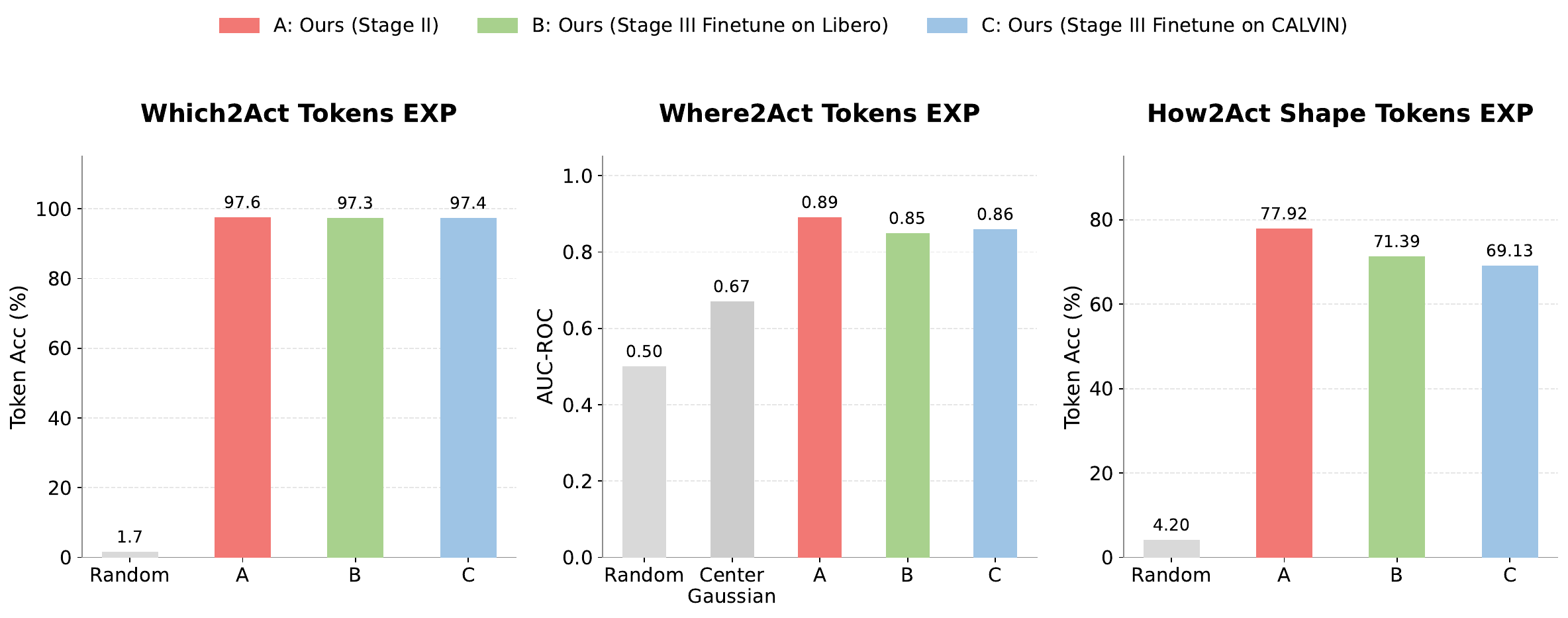}
        \caption{Quantitative Results for Subgoal Tokens.}
        \label{fig:subgoal_token_acc}
    \end{subfigure}
    
    \begin{subfigure}{\linewidth}
        \centering
        \includegraphics[width=\linewidth]{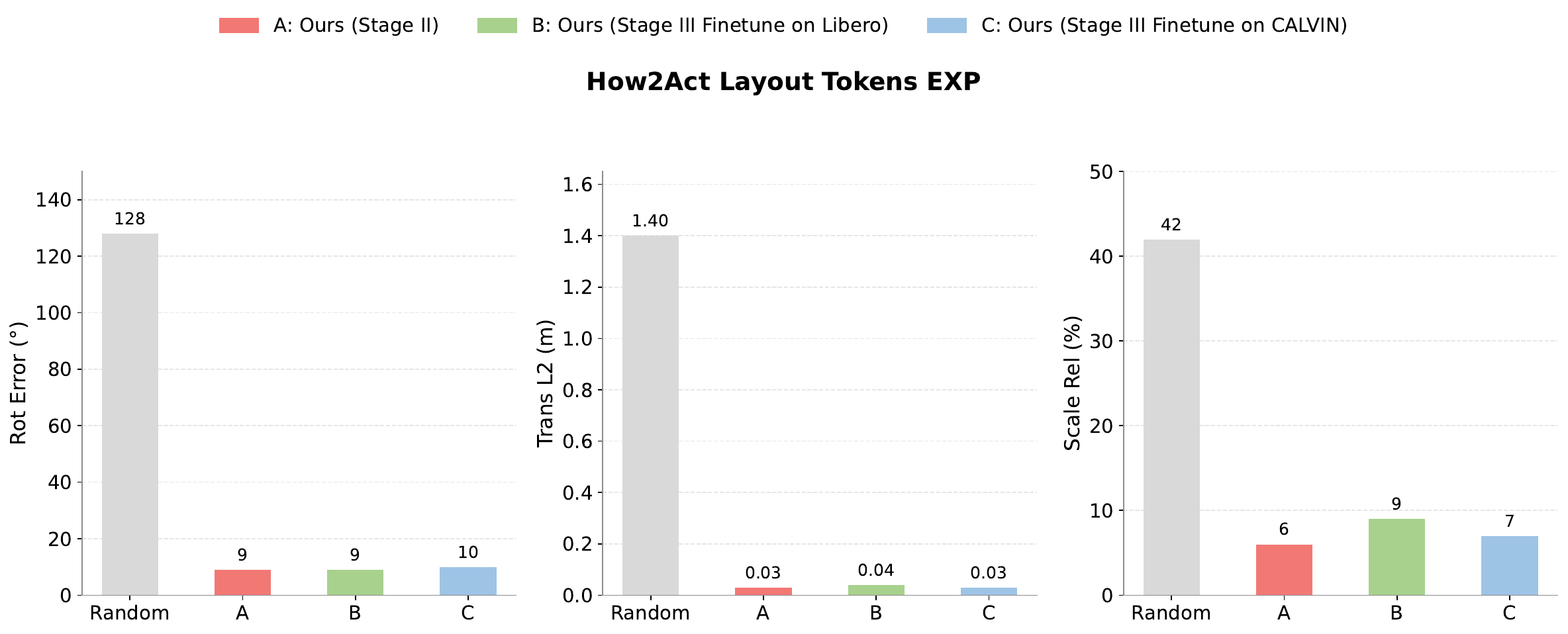}
        \caption{Quantitative Results for How2Act Layout Tokens.}
        \label{fig:subgoal_layout}
    \end{subfigure}

    \caption{\textbf{Prev-Variant Quantitative Diagnostics.} The discrete VQ-VAE variant captures task-relevant semantics and 3D spatial information across its original metrics.}
    \label{fig:combined_results}
\end{figure}

For the Prev variant, Fig.~\ref{fig:subgoal_token_acc} and Fig.~\ref{fig:subgoal_layout} show that all three affordance query tokens extract crucial environmental context and substantially outperform random baselines. Which2Act and Where2Act achieve near-perfect performance, while the lower absolute accuracy of the How2Act Shape token reflects the difficulty of reconstructing high-fidelity 3D voxels from highly compressed VLA tokens.

However, this limitation does not impede downstream control. The generated 3D representations sufficiently capture essential task progression characteristics—such as the coarse physical structure of the manipulated object and the anticipated interaction modality—serving as effective explicit planning targets for the Action Expert. Furthermore, our \textbf{extra ablation studies} reveal that without Which2Act and Where2Act, the model experiences a severe performance degradation, with the LIBERO average success rate dropping to $11.8\%$. This underscores that Which2Act and Where2Act provide critical implicit object-centric priors (functioning similarly to bounding boxes or segmentation masks in traditional explicit 3D models) that facilitate the learning of How2Act. Conversely, the successful convergence of How2Act further corroborates that Which2Act and Where2Act have indeed learned robust grounding capabilities. This reciprocal relationship strongly validates our architectural design, demonstrating that the three tokens operate synergistically to comprehend the manipulation scene.

\subsection{Representation Decoupling: Backbone vs. Decoder}
\label{sec:subgoals_decoupling}

To confirm that the subgoals genuinely facilitate action generation rather than merely overfitting a high-capacity decoder, it is imperative to verify that the subgoal losses primarily optimize the backbone representations. To this end, we design a decoupling experiment: we freeze our Stage II backbone (trained for 100k steps) and evaluate its feature extraction quality using decoders trained with varying, lower step counts (from 5k to 100k steps).

\begin{figure}[t]
    \centering
    \includegraphics[width=\linewidth]{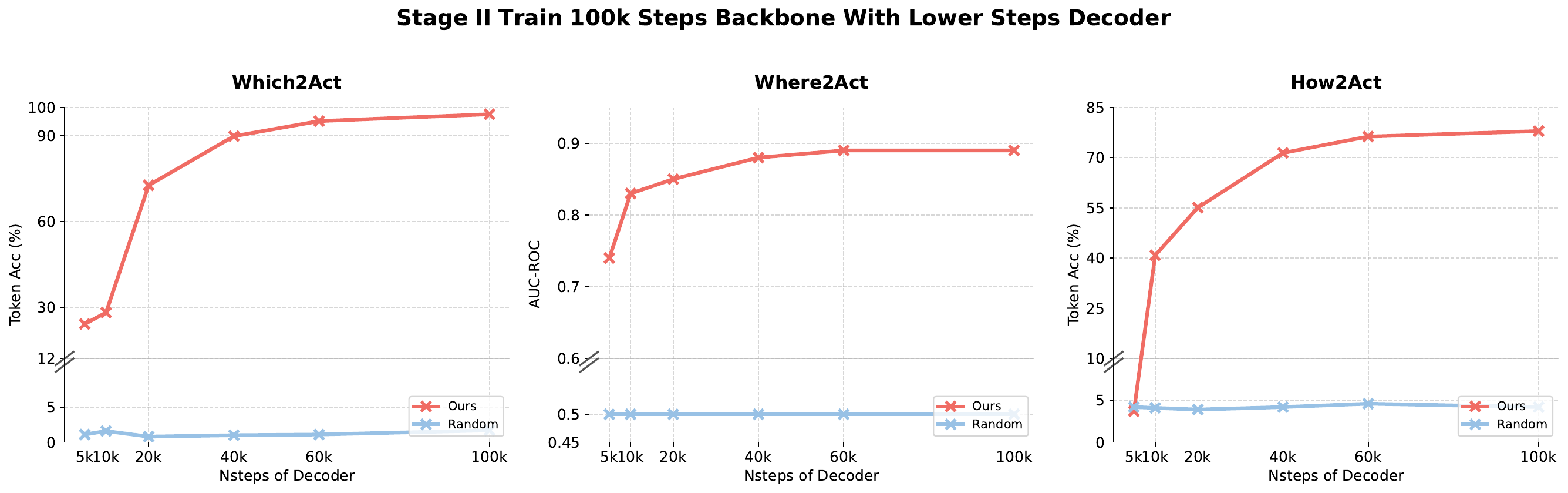}
    \caption{\textbf{Backbone Representation Evaluation.} Performance of a fully trained VLA backbone (100k steps) when evaluated with decoders trained for fewer steps (5k to 100k). The backbone maintains robust feature extraction even when paired with a weakly trained decoder.}
    \label{fig:three_decoder}
\end{figure}

As shown in Fig.~\ref{fig:three_decoder}, even when paired with an under-trained decoder, the fully trained backbone still extracts highly meaningful affordance features. The performance exhibits a stable, monotonic increase as the decoder steps align with the backbone, rather than catastrophically failing under the weak decoder's evaluation.
This confirms that the VLA backbone itself intrinsically assimilates the affordance representations to guide actions.

\subsection{Qualitative Analysis of Affordance Grounding}
\label{sec:subgoals_qualitative}

\begin{figure}[t]
    \centering
    \includegraphics[width=\linewidth]{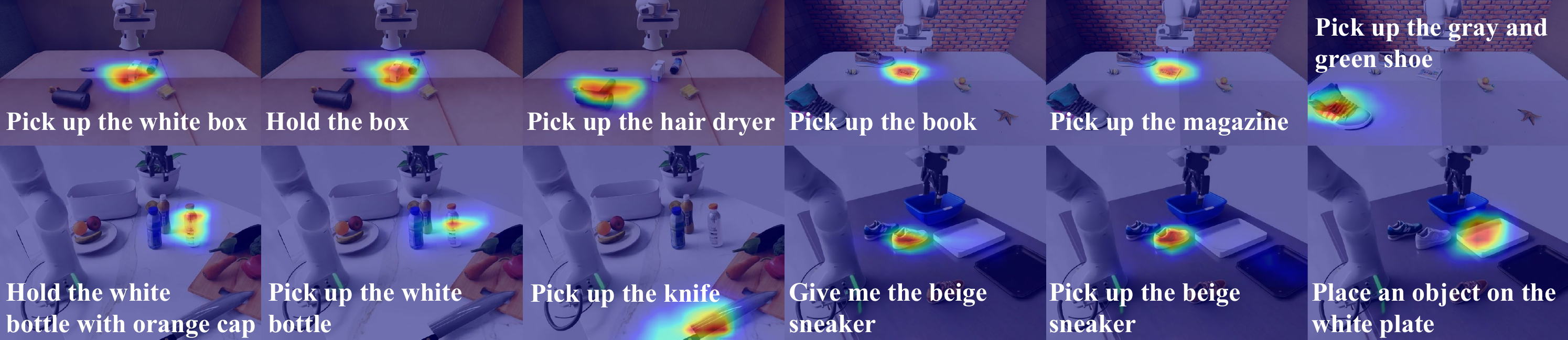}
    \caption{\textbf{Qualitative Visualization of Where2Act Affordance Maps.} The model demonstrates robust Vision-Language alignment, dynamically adjusting its affordance predictions based on varying language instructions given the identical visual observation.}
    \label{fig:qualitative_affordance}
\end{figure}

To intuitively illustrate the model's grounding capabilities, we visualize the Where2Act Affordance Maps in Fig.~\ref{fig:qualitative_affordance}.

We specifically design a Vision-Language (V-L) bridging experiment: given an identical visual observation, we alter the language instruction ($L$). The generated affordance map dynamically adapts to $L$, indicating that our model deeply comprehends V-L alignment rather than merely relying on visual salience biases. Furthermore, the model exhibits strong robustness; semantically similar commands consistently yield highly correlated affordance heatmaps, whereas divergent commands elicit clean and precise shifts in the affordance focus.

\section{Model Variants and Design Choice Details}
\label{sec:supp_arch}
\subsection{Design Insight}
\label{sec:design_insight}

A central \textbf{insight} of this work is that \emph{blindly scaling up data fails to maximize the intrinsic power within the datasets}. Through extensive experiments, we observe that Vision-Language-Action (VLA) models are fundamentally representation learning systems---their performance ceiling is jointly determined by the quality of both the learned representations and the training data, not merely the quantity of data. This insight drives our design of AffordanceVLA, which centers on a conceptual bridge supported by two key technical design details:
\begin{enumerate}
\item \textbf{Affordance as an intermediate bridge:} At the conceptual level, we introduce affordance as a structural bridge to link visual perception with robotic execution. 
\item \textbf{Design detail I: Architectural refinement:} To support high-fidelity affordance representations, we transition from a discrete VQ-VAE to a continuous Flux VAE for Which2Act supervision. This shift eliminates codebook quantization artifacts and enables the capture of finer-grained visual cues essential for precise manipulation.
\item \textbf{Design detail II: Data quality investment:} To ensure the architectural capacity is grounded in accurate knowledge, we curate high-quality datasets (InternData-A1 and DROID) and implement a rigorous automated annotation pipeline. This ensures near-perfect accuracy in affordance labels.
\end{enumerate}

The synergy of this bridge-centric architecture and its supporting technical details is what enables AffordanceVLA to break through performance ceilings that data scaling alone cannot overcome (as validated by the data-efficiency analysis in the main text, \Cref{sec:ablation}).

\subsection{From VQ-VAE to Flux VAE: Which2Act Redesign}
\label{sec:which2act_redesign}

In our preliminary version (denoted Prev), \textbf{Which2Act} supervision relied on a frozen multi-scale VQ-VAE encoder.
The VQ-VAE's discrete codebook structure imposes a hard constraint on the token count $N_{w2l}$: specifically, $\sqrt{N_{w2l}}$ must lie within the predefined scale list $\{1,2,3,4,5,6,8,10,13,16\}$.
To achieve sufficient spatial resolution, we set $\sqrt{N_{w2l}} = 8$, requiring $N_{w2l} = 64$ tokens.
However, with $N = 64$ tokens and a fixed total latent dimension of 16{,}384 floats (from the $256 \times 256$ input), each token's projection target is only $D_{\text{proj}} = 16{,}384 / 64 = 256$ dimensions---far below the Generation Expert's hidden size $D_{\text{gen}} = 1024$.
This forces the projection head to compress each token by $4\times$, squandering the expert's representational capacity.

In AffordanceVLA, we replace VQ-VAE with a frozen Flux VAE encoder that produces continuous latents $z_q \in \mathbb{R}^{B \times 16 \times 32 \times 32}$ (16{,}384 floats total; 16 channels with $8\times$ spatial downsampling of the $256 \times 256$ crop).
With continuous targets and MSE supervision, the token count $N_{w2l}$ is freed from codebook constraints.
We identify $N_{w2l} = 16$ as one choice because:

\begin{equation}
D_{\text{proj}} = \frac{16{,}384}{N_{w2l}} = \frac{16{,}384}{16} = 1024 = D_{\text{gen}}
\end{equation}

At this configuration, the projection head performs an equal-dimension nonlinear transform ($D_{\text{gen}} \to D_{\text{proj}}$) with zero information compression or expansion, maximizing the per-token representational utility.
Increasing the token count merely fragments the same information across more tokens while inflating the shared attention cost; decreasing it forces each token to carry a larger projection burden, requiring lossy expansion.

\subsection{Model Variants}
\label{sec:model_variants}

We have three model configurations, summarized in \Cref{tab:model_variants}.
At each scale, an optional set of wrist tokens—which shares the same decoder architecture and weights as Which2Act—can be appended, supervised by the full wrist camera image.
In practice, the wrist token yields marginal improvement and occasionally weakens performance; we therefore treat it as optional and exclude it from the main results.

\begin{table}[h]
    \centering
    \caption{
      \textbf{AffordanceVLA model variants.}
      $N_{\text{gen}}$ is the total number of Affordance Generation expert tokens (excluding the optional Wrist group).
      Prev uses a frozen VQ-VAE with discrete codebook supervision (CrossEntropy), requiring $\sqrt{N_{w2l}} \in \{1,...,16\}$ (hard constraint).
      AffordanceVLA-fast and AffordanceVLA use a frozen Flux VAE with continuous latent supervision (MSE), freeing the token count from codebook constraints.
    }
    \label{tab:model_variants}
    \begin{tabular}{lcccccc}
        \toprule
        \multirow{2}{*}{\textbf{Variant}} & \multirow{2}{*}{\textbf{Which2Act}} & \multirow{2}{*}{\textbf{Where2Act}} & \multicolumn{2}{c}{\textbf{How2Act}} & \multirow{2}{*}{\textbf{$N_{\text{gen}}$}} & \textbf{Which2Act} \\
        \cmidrule(lr){4-5}
        & & & Shape & Layout & & \textbf{Encoder} \\
        \midrule
        Prev            & 64 & 16  & 30  & 2 & 112 & VQ-VAE \\
        AffordanceVLA-fast & 16 & 16  & 30  & 2 & 64  & Flux VAE \\
        AffordanceVLA      & 64 & 64  & 256 & 4 & 388 & Flux VAE \\
        \bottomrule
    \end{tabular}
\end{table}

\paragraph{Token allocation analysis.}
For \textbf{Where2Act}, tokens are decoded into a heatmap via Transformer cross-attention. For \textbf{How2Act Shape}, the decoder must condition a 3D diffusion process to reconstruct a $16^3$ voxel latent. The conditioning pathway applies mean-pooling over shape tokens.
While this bottleneck limits instance-level 3D fidelity, it suffices for encoding categorical shape priors (``cup-shaped'' vs. ``plate-shaped'') that meaningfully guide downstream action generation.
The AffordanceVLA variant allocates 256 shape tokens, providing a richer upstream representation that the mean-pooling aggregates more informatively.
\textbf{How2Act Layout} regresses a compact 10D vector (rotation$_4$ + scale$_3$ + translation$_3$); even 2 or 4 tokens with mean-pooling are sufficient.

\section{Data-Centric Methodology}
\label{sec:supp_data}

\subsection{Data Quality as the Performance Ceiling}
\label{sec:supp_data_quality}

Since VLA is fundamentally representation learning, the quality of training data directly determines the model's performance ceiling: if the supervision signals are noisy or inconsistent, the learned representations will be fundamentally flawed regardless of model capacity.
This principle guides our dataset selection and annotation strategy:

\paragraph{Dataset selection.}
For Stage~II co-training, we only use the full InternData-A1 simulation dataset, renowned for its photorealistic pixel-level rendering quality across diverse manipulation tasks (basic, pick-and-place, and long-horizon categories).
For real-world post-training, we select a curated subset of the DROID dataset, a large-scale in-the-wild robot manipulation corpus collected across diverse real-world environments with consistent high-quality demonstrations.
Building upon the Stage II co-training, this phase serves as a critical intermediate step designed to enhance the model's generalization in diverse real-world settings before proceeding to final In-house trajectory fine-tuning. By exposing the model to the rich physical priors in DROID, we effectively bridge the sim-to-real gap, ensuring a more robust transition from large-scale pre-training to specific downstream tasks.

\paragraph{Annotation quality guarantee.}
Crucially, our affordance annotation pipeline (detailed in \Cref{sec:supp_pipeline}) achieves near-perfect accuracy on both A1 and DROID, precisely because these datasets provide clean, high-resolution observations that enable reliable visual grounding.
This creates a positive propagative effect: high-quality source data $\to$ high-quality annotations $\to$ high-quality representations $\to$ superior downstream performance.

\subsection{Affordance Annotation Pipeline}
\label{sec:supp_pipeline}

Stage~II (A1) and Stage~III (LIBERO, CALVIN, DROID) robot datasets do not natively contain affordance annotations (bbox, heatmap, shape token, layout token).
We design a fully automated annotation pipeline to generate high-quality affordance supervision for these datasets.
Stage~I supervision follows the Grounding and Pointing pools in \Cref{tab:stage1_data_pool}. Native spatial annotations are normalized into common Which2Act and Where2Act targets. When a source provides a verified interaction point but no dense heatmap, we use SAM within its valid object region and construct a Gaussian heatmap centered at that point. All retained samples are further augmented with SAM-3D-generated shape and layout targets.

\subsubsection{Step 0: RexOmni Fine-Tuning.}
While the open-source visual grounding model RexOmni natively possesses robust, general-purpose object detection and spatial pointing capabilities, we further adapt it on the embodied-manipulation portion of the Stage~I Pointing pool (\Cref{tab:stage1_data_pool}). Rather than learning these concepts from scratch, this adaptation aligns RexOmni's visual grounding priors with the geometric and physical requirements of robotic affordance localization.
Specifically, it enhances the model's ability to produce geometrically consistent outputs: the bounding box identifying the target object and the key interaction point indicating where to act.


\subsubsection{Step 1: Rule-Based Keyframe Detection.}
For each trajectory, we extract semantically meaningful keyframes from the robot state signal (joint positions, gripper states) using six complementary detection rules, inspired by BridgeVLA:
\begin{itemize}
    \item \textbf{Start:} The first frame of the trajectory, capturing the initial scene configuration.
    \item \textbf{Pre-Action:} $N$ frames before each gripper state change ($N{=}30$ by default). This ``approach phase'' captures the visual observation most indicative of where to act, as the robot is oriented toward but has not yet contacted the target.
    \item \textbf{Gripper:} The exact frame where the gripper's open/close state flips, marking the precise grasp or release event.
    \item \textbf{Stop:} Frames where all joint velocities fall below threshold ($\|\dot{q}_j\| < \epsilon$, $\epsilon{=}0.01$) while the gripper state remains stable across neighboring frames. These correspond to mid-trajectory pauses, typically at sub-task transitions (e.g., stabilization after placement).
    \item \textbf{Apex:} Frames where the joint velocity norm is a local minimum with $\epsilon < \|\dot{q}\| < 5\epsilon$. These capture trajectory turning points where the motion direction reverses (e.g., transitioning from approach to retraction).
    \item \textbf{End:} The frame $M$ steps before the trajectory end ($M{=}25$), avoiding potentially unstable or static trailing frames.
\end{itemize}

\noindent
All detected keyframes are sorted chronologically, and pairs separated by fewer than 10 frames are merged (retaining the later index and aggregating reason labels) to eliminate redundancy.

\subsubsection{Step 2a: Instruction Decomposition via a Text LLM (Claude Opus~4.5).}
A long-horizon command (\eg ``Pick up the red cup and put it on the shelf'') spans multiple primitive interactions, each aligned to a different keyframe. We therefore first decompose it using a text-only large language model, Claude Opus~4.5. The LLM receives the \emph{original instruction} together with the \emph{ordered keyframe-meaning sequence} from Step~1 (each keyframe tagged by its semantic role, \eg \textit{Pre-Action}, \textit{Gripper}, \textit{Stop}), and returns one atomic sub-instruction per keyframe that is temporally consistent with the keyframe roles. The exact template is shown in \textbf{Prompt~A}.

\begin{promptbox}{Prompt A --- Instruction Decomposition (Claude Opus~4.5, text-only LLM)}
\textbf{Role.} You are an expert robotic-manipulation annotator.

\textbf{Task.} Decompose a long-horizon instruction into atomic, single-interaction sub-instructions---\emph{one per keyframe}---that together accomplish the task. Each sub-instruction must describe exactly one primitive interaction (a single grasp, place, push, or release), respect the temporal order and the semantic role of its keyframe, and introduce no step that is not implied by the original instruction.

\textbf{Inputs.}\\
$\bullet$~Original instruction: \texttt{[original\_instruction]}\\
$\bullet$~Keyframe-meaning sequence: \texttt{[(index, role), ...]}

\textbf{Output (strict JSON).} A list aligned to the keyframes; each item is \texttt{\{"keyframe\_index": [int], "sub\_instruction": [str]\}}.
\end{promptbox}

\subsubsection{Step 2b: Per-Keyframe Affordance Annotation via a VLM (Qwen3-VL).}
Given the per-keyframe sub-instructions, we then query our deployed \textbf{Qwen3-VL-235B}~\cite{qwen3} multimodal model on each keyframe \emph{image} to emit the two expressions that drive grounding. To remain compatible with the downstream grounding model, the prompt follows RexOmni~\cite{rexomni} conventions: concise category names for detection and ``where-to'' referring expressions for pointing. Crucially, the VLM is conditioned on the full context---the \emph{original instruction}, the \emph{keyframe-meaning sequence}, the \emph{current keyframe index}, the \emph{current keyframe meaning}, and the \emph{sub-instruction} from Step~2a---which disambiguates the intended target far more reliably than the keyframe image alone. The template is shown in \textbf{Prompt~B}, and it produces:
\begin{enumerate}
    \item \textbf{Detection category:} the single most specific target the gripper directly contacts at this keyframe (\eg ``drawer handle'' rather than ``drawer'' for grasping, ``shelf surface'' for placing, or ``button'' for pushing), focusing on the localized interacting region.
    \item \textbf{Affordance instruction:} a spatial ``where'' reformulation of the action intent (\eg ``Where to grasp the drawer handle to open it?'', ``Where to place the cup securely on the shelf?'', or ``Where to push the button?'').
\end{enumerate}
These two expressions respectively drive the Which2Act (\emph{what object}) and Where2Act (\emph{where to interact}) supervision signals.

\begin{promptbox}{Prompt B --- Per-Keyframe Affordance Annotation (Qwen3-VL, RexOmni-style)}
\textbf{Role.} You prepare prompts for RexOmni, a detection-and-pointing model driven by category names and referring expressions.

\textbf{Task.} For the current keyframe, output (i) the single most specific target part the gripper directly interacts with, and (ii) a spatial ``where-to'' affordance query for that interaction. Follow RexOmni conventions: the detection category is a concise noun phrase (\eg ``drawer handle'', not ``drawer''); the affordance instruction is a referring expression of the form ``Where to [verb] the [target] to [goal]?''.

\textbf{Inputs.}\\
$\bullet$~Original instruction: \texttt{[original\_instruction]}\\
$\bullet$~Keyframe-meaning sequence: \texttt{[(index, role), ...]}\\
$\bullet$~Current keyframe index: \texttt{[int]}\\
$\bullet$~Current keyframe meaning: \texttt{[role]}\\
$\bullet$~Sub-instruction: \texttt{[sub\_instruction]}\\
$\bullet$~Current keyframe image: \texttt{[image]}

\textbf{Output (strict JSON).} \texttt{\{"detection\_category": [str], "affordance\_instruction": [str]\}}.
\end{promptbox}

\noindent\textbf{Note.} The prompts shown above are illustrative examples; the complete dataset-specific prompt templates are implemented in the data-preprocessing module of our codebase, and the resulting preprocessed annotations will also be made publicly available.

\subsubsection{Step 3: Visual Grounding and Affordance Generation.}
The two instruction types from Step 2 are fed to the fine-tuned RexOmni in dual modes: 
\textit{detection categories} $\to$ \textit{detection mode} $\to$ bbox; 
\textit{affordance instructions} $\to$ \textit{pointing mode} $\to$ affordance point. 
From the resulting bbox and point, we generate the full affordance annotation:

\begin{enumerate}
    \item \textbf{Visual latent (Which2Act GT):} We crop the observation according to the target bbox and extract a continuous visual latent.
    \item \textbf{Heatmap (Where2Act GT):} We run SAM within the bbox region to obtain a fine-grained segmentation mask, then generate a Gaussian affordance heatmap centered at the affordance point and constrained by the mask boundary. 
    \item \textbf{Shape token \& Layout token (How2Act GT):} The bbox crop and SAM mask are input to a SAM-3D service, which produces a shape token $\in \mathbb{R}^{4096 \times 8}$ (3D voxel latent) and a layout token $\in \mathbb{R}^{10}$ (rotation$_4$ + scale$_3$ + translation$_3$). 
\end{enumerate}

\subsubsection{Step 4: Rigorous Quality Verification.}
We implement multi-level quality control to ensure annotation reliability:
\begin{enumerate}
    \item \textbf{Pipeline consistency --- Point-in-Bbox verification:} After running the full pipeline, we verify that 100\% of generated affordance points fall strictly within their corresponding bounding boxes.
    Only when this spatial consistency check passes do we certify the pipeline as operationally correct, fundamentally ensuring that the detection (bbox) and pointing (point) outputs are geometrically coherent.
    Successfully passing this rigorous geometric coherence check not only certifies the operational correctness of the pipeline but also serves as the definitive standard for verifying the RexOmni fine-tuning, ensuring that its decoupled detection (bbox) and pointing (point) outputs are perfectly aligned for downstream manipulation tasks.
    \item \textbf{Human audit --- 100-round random sampling:} We conduct 100 independent random sampling rounds, each drawing 30 annotation samples for manual inspection.
    Only when all 100 rounds (totaling 3{,}000 inspected samples) achieve a 100\% pass rate is the batch admitted into model training.
    This stringent audit protocol eliminates systematic errors and edge cases, providing strong empirical evidence of annotation quality.
\end{enumerate}

\section{Dataset Details}
\label{sec:data_statistics}
\Cref{tab:data_statistics} summarizes all datasets across training stages.

\begin{table}[h]
    \centering
    \resizebox{\linewidth}{!}{
        \begin{tabular}{llllllll}
            \toprule
                \textbf{Dataset} & \textbf{Source} & \textbf{Stage} & \textbf{Embodiment} & \textbf{FPS} & \textbf{\# Samples} & \textbf{Type} \\
            \midrule
                Curated Grounding $+$ Pointing Pool & Mixed & I & Mixed / -- & -- & 8.8M & Spatial supervision \\
            \midrule
                A1 (InternData) & In-house       & II      & Franka    & 30  & 149K     & Trajectory \\
            \midrule
                LIBERO          & Public         & III     & Franka    & 20  & --     & Trajectory \\
                CALVIN          & Public         & III     & Franka    & 30  & -- & Trajectory \\
            \midrule
                DROID (subset)  & Public         & Real    & Franka     & 15  & $\sim$150K   & Trajectory \\
                Close microwave   & In-house       & Real    & Franka   & 15  & 80   & Trajectory \\
                Close safe   & In-house       & Real    & Franka   & 15  & 80   & Trajectory \\
                Pick up red cup   & In-house       & Real    & Franka   & 15  & 80   & Trajectory \\
                Pick up green cup   & In-house       & Real    & Franka   & 15  & 80   & Trajectory \\
                Pick up duck   & In-house       & Real    & Franka   & 15  & 100   & Trajectory \\
                Pick up banana   & In-house       & Real    & Franka   & 15  & 80   & Trajectory \\
                Pick up flower   & In-house       & Real    & Franka   & 15  & 100   & Trajectory \\
                Pick up bear   & In-house       & Real    & Franka   & 15  & 100   & Trajectory \\
                Pick The Item in Drawer   & In-house       & Real    & Franka   & 15  & 100   & Trajectory \\
                Close the drawer   & In-house       & Real    & Franka   & 15  & 100   & Trajectory \\
                Pick the bread   & In-house       & Real    & Franka   & 15  & 100   & Trajectory \\
                Toast the bread   & In-house       & Real    & Franka   & 15  & 150   & Trajectory \\
                Pick all the rubbish   & In-house       & Real    & Franka   & 15  & 200   & Trajectory \\
            \bottomrule
        \end{tabular}
    }
    \caption{
      \textbf{Data statistics.}
      Stage~I source composition is detailed in \Cref{tab:stage1_data_pool}.
      Stage~II uses the full A1 simulation dataset exclusively.
      Stage~III fine-tunes on the target benchmark (LIBERO or CALVIN).
      Robot data are augmented with automatically generated affordance annotations via our pipeline (\Cref{sec:supp_pipeline}); Stage~I data are additionally augmented with SAM-3D-generated shape and layout tokens. The full command for ``Toast the bread'' is ``Toast the bread with toaster button''.
    }
    \label{tab:data_statistics}
\end{table}

\section{Training Details}
\label{sec:training_recipe}

\begin{table}[h]
    \centering
    \resizebox{\linewidth}{!}{
        \begin{tabular}{lllllll}
            \toprule
                \multirow{2}[1]{*}{\textbf{Hyperparameters}} & \multirow{2}[1]{*}{\textbf{Stage I}} & \multirow{2}[1]{*}{\textbf{Stage II}} & \multicolumn{3}{c}{\textbf{Stage III}} & \multirow{2}[1]{*}{\textbf{Real-World}} \\
                \cmidrule{4-6}
                & & & \textbf{LIBERO} & \textbf{CALVIN} & \textbf{DROID} & \\
            \midrule
                Batch Size (effective)        & 3{,}072  & 1024        & 256    & 256    & 1024    & 128 \\
                Learning Rate                 & $5 \times 10^{-5}$ & $5 \times 10^{-5}$ & $5 \times 10^{-5}$ & $5 \times 10^{-5}$ & $5 \times 10^{-5}$ & $5 \times 10^{-5}$ \\
                LR Scheduler                  & Cosine   & Constant     & Cosine & Cosine & Constant & Cosine \\
                Loss Weight (Afd : Act)       & -- & 0.5 : 1     & 0.15 : 1 & 0.15 : 1 & 0.15 : 1 & 0.15 : 1 \\
                Training Steps                & 300K     & 230K       & --   & --   & 200K     & -- \\
                Training Epochs               & --       & --         & 20     & 20     & --     & 40 \\
            \midrule
                Input Resolution              & $224 \times 224$ & $224 \times 224$ & $224 \times 224$ & $224 \times 224$ & $224 \times 224$ & $224 \times 224$ \\
                Which2Act Crop Size           & $256 \times 256$ & $256 \times 256$ & $256 \times 256$ & $256 \times 256$ & $256 \times 256$ & $256 \times 256$ \\
                Action Chunk Size             & --       & 30         & 6     & 6     & 30     & 50 \\
                Denoise Steps                 & --       & 10         & 10     & 10     & 10     & 10 \\
            \bottomrule
        \end{tabular}
    }
    \caption{
      \textbf{Training recipe of AffordanceVLA} on 16$\times$ NVIDIA H200 GPUs.
      The ``Afd : Act'' row reports the \emph{task-level} weight ratio $\lambda_{\text{afd}}{:}\lambda_{\text{act}}$ between the aggregated affordance loss (``Afd'') and the action flow-matching loss (``Act''), with $\lambda_{\text{act}}$ normalized to $1$. The internal weighting among the four affordance sub-objectives is held fixed across all stages at $\lambda_{\text{which}}{:}\lambda_{\text{where}}{:}\lambda_{\text{shape}}{:}\lambda_{\text{layout}} = 0.1{:}0.1{:}0.1{:}0.04$ (equivalently $5{:}5{:}5{:}2$).
      In Stage~I, the action loss is inactive (no ground-truth actions).
    }
    \label{tab:training_recipe}
\end{table}

All training is conducted on a cluster of 16$\times$ NVIDIA H200 GPUs (141\,GB HBM3e per card).
\Cref{tab:training_recipe} summarizes the hyperparameters across all stages.
In Stage~I, only the Affordance Generation expert, Affordance Query, and all decoders are trained; the Understanding and Action experts remain frozen.
In Stage~II and~III, all experts and decoders are jointly trained end-to-end, with the vision encoder fine-tuned at a lower learning rate.
The Pi0 and Pi0.5 results on LIBERO-Plus in \Cref{tab:libero_plus} are measured under our evaluation setup and therefore differ from the official leaderboard.

\section{Inference Latency}
\label{sec:latency_breakdown}

\Cref{tab:latency_breakdown} reports the inference latency breakdown for the AffordanceVLA model deployed on an NVIDIA RTX 5090 GPU (32\,GB GDDR7).

\begin{table}[h]
    \centering
    \begin{tabular}{l|c}
        \toprule
            \textbf{Component} & \textbf{Latency (ms)} \\
        \midrule
            Image preprocessing (resize $+$ normalize) & $\sim$6 \\
            Phase 1: Image encoder (SigLIP) & $\sim$22 \\
            Phase 1: Understanding $+$ Affordance Generation & $\sim$52 \\
            Phase 2: $\times$10 Action denoising (Euler flow matching) & $\sim$92 \\
            Misc (tokenization, state preparation) & $\sim$4 \\
        \midrule
            \textbf{Total inference} & $\sim$\textbf{176} \\
        \bottomrule
    \end{tabular}
    \caption{
      \textbf{Inference latency of AffordanceVLA on RTX 5090.}
      The total inference time of $\sim$176\,ms enables real-time control at $\sim$5.7\,Hz.
      The robot is connected via wired Ethernet; reported numbers exclude communication delays.
    }
    
    \label{tab:latency_breakdown}
\end{table}


\section{Real-World Experiments Details and Extra Experiments}
\label{sec:real}

Our real-world experiments are conducted on a 7-DoF Franka Emika Panda manipulator equipped with a DH-Robotics PGC-140 parallel jaw gripper (50\,mm stroke, 140\,N grip force). Visual observations are captured by two Intel RealSense D435 cameras (one fixed third-person view, one wrist-mounted) streaming RGB images at 15\,Hz. The robot connects to the host workstation via wired Ethernet.

To bridge the sim-to-real gap, the model is first pre-trained on a curated subset of the DROID dataset, then fine-tuned on in-house demonstrations. During deployment, the model processes single-frame observations, executing the first 5 steps from each predicted action chunk before re-planning.

\begin{figure}[htbp]
    \centering
    \includegraphics[width=\textwidth]{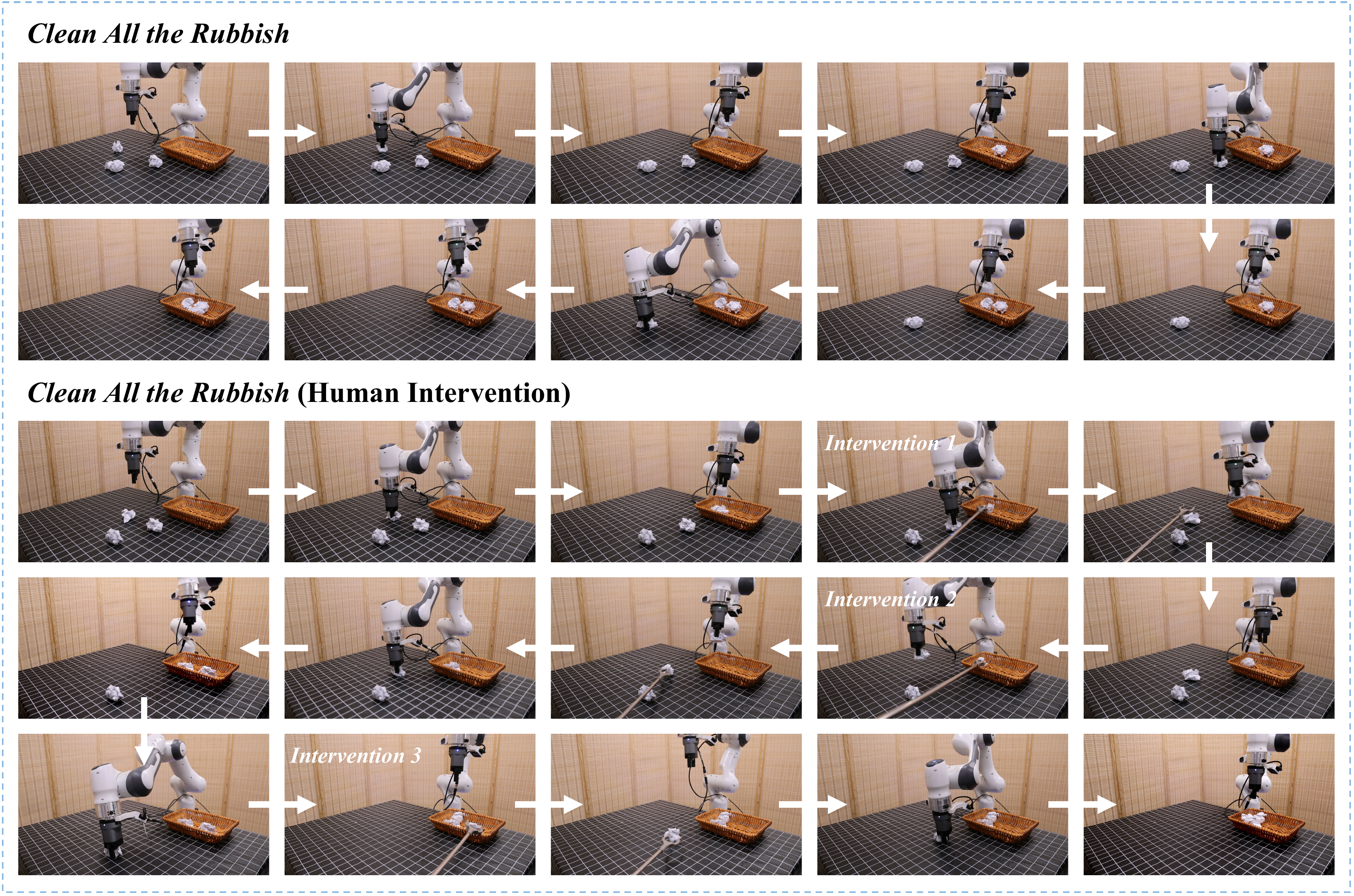}
    \caption{Sequential execution of the ``Clean all the rubbish'' task, alongside a demonstration of human intervention. The policy's ability to recover from external disturbances highlights its strong generalizability and robustness.}
    \label{fig:real_supp2}
\end{figure}

To comprehensively evaluate our method, we design a diverse set of manipulation tasks. To highlight the generalizability and robustness of our learned policy, we present a more complex task, ``Clean all the rubbish'', as shown in Figure~\ref{fig:real_supp2}. Notably, this figure also features a trial with explicit human intervention. The successful recovery from dynamic external disturbances demonstrates that our closed-loop re-planning strategy enables the robot to seamlessly adapt to unexpected changes in the environment.



\end{document}